\documentclass{article}

\usepackage[main, final]{neurips_2026}

\usepackage[utf8]{inputenc} %
\usepackage[T1]{fontenc}    %
\usepackage{hyperref}       %
\usepackage{url}            %
\usepackage{booktabs}       %
\usepackage{amsfonts}       %
\usepackage{nicefrac}       %
\usepackage{microtype}      %
\usepackage{xcolor}         %

\usepackage{url}
\usepackage{xspace}
\usepackage{graphicx}
\usepackage{wrapfig}
\usepackage{stmaryrd}
\usepackage{amssymb}
\usepackage{enumitem}
\usepackage[most]{tcolorbox} %
\usepackage{caption}
\usepackage{multirow}
\usepackage{rotating}
\usepackage{microtype}
\usepackage{graphicx}
\usepackage{subcaption}
\usepackage{booktabs} %
\usepackage{xcolor}      %
\usepackage{pifont}      %

\usepackage{hyperref}

\usepackage{amsmath}
\usepackage{amssymb}
\usepackage{mathtools}
\usepackage{amsthm}
\usepackage{float}

\usepackage[capitalize,noabbrev]{cleveref}

\theoremstyle{plain}

\theoremstyle{definition}

\theoremstyle{remark}

\usepackage[disable,textsize=tiny]{todonotes}

\newcommand{\DATANAME}{\textsc{MathLens}\xspace}
\newcommand{\DATANAMEG}{\textsc{MathLens-NS}\xspace}

\definecolor{arrowgreen}{HTML}{4E6E4F}  %
\definecolor{arrowred}{HTML}{E64545}  %

\tcbset{
  findingboxstyle/.style={
    colback=gray!10,        %
    colframe=black,       %
    boxrule=1pt,        %
    arc=3pt,              %
    left=4pt, right=4pt,  %
    top=4pt, bottom=4pt,  %
  }
}

\title{What MLLMs Learn about When they Learn about Multimodal Reasoning}

\author{%
  Jiwan Chung$^{1,2}$\thanks{Work done while interning at Microsoft Research}\;,
  Neel Joshi$^1$\;,
  Pratyusha Sharma$^{4}$\thanks{Work done while at Microsoft Research}\;,
  Youngjae Yu$^3$\thanks{co-principal investigators}\;,
  Vibhav Vineet$^{1\ddagger}$ \\
  \\
  $^1$Microsoft Research AI Frontiers, $^2$Yonsei University\\
  $^3$Seoul National University, $^4$New York University\\
  \texttt{jiwan.chung.research@gmail.com, vivineet@microsoft.com}\\
  Project page: \url{https://jiwanchung.github.io/mathlens/}\\
}

\begin{document}

\maketitle

\begin{abstract}
Evaluation of multimodal reasoning models is typically reduced to a single accuracy score, implicitly treating reasoning as a unitary capability. We introduce \DATANAME, a benchmark of textbook-style geometry problems that exposes this assumption by operationally decomposing performance into perception, reasoning, and multimodal-specific components. Each problem is derived from a symbolic specification and accompanied by visual diagrams, text-only variants, multimodal questions, and targeted perceptual probes, enabling controlled measurement of each component. Using this decomposition, we show that common training strategies are associated with systematically different capability profiles that are invisible under aggregate accuracy. Across the models we study, reinforcement learning primarily improves perceptual grounding and robustness to diagram variation, while our analysis suggests that textual SFT yields gains through reflective reasoning. In contrast, as perception and reasoning improve, a growing fraction of remaining errors fall outside these components and are categorized as multimodal-specific. These results suggest that apparent progress in multimodal reasoning reflects shifting balances among subskills rather than uniform advancement, motivating evaluation beyond scalar accuracy.
\end{abstract}

\section{Introduction}
\label{sec:intro}

Recent advances in reasoning with Large Language Models (LLMs) have yielded remarkable progress in challenging domains such as Olympiad-level mathematics~(\cite{aime2025}), graduate-level scientific question answering~(\cite{rein2024gpqa}), and multi-step program synthesis~(\cite{austin2021program,chen2021evaluating}).
Motivated by these successes, a natural extension is to adapt similar training paradigms to Multimodal Large Language Models (MLLMs), equipping them with reasoning capabilities over both text and visual inputs. Tasks such as mathematical problem solving~(\cite{lu2024mathvista,wang2024mathvision}) and visual puzzles~(\cite{dao2025alphamaze,ghosal2024language,feng2025visualsphinx}) illustrate the potential of this direction, giving rise to methods that adapt Supervised FineTuning (SFT)~(\cite{sun2025mitigating,chung2025don}) and Reinforcement Learning (RL)~(\cite{deng2025openvlthinker,meng2025mmeureka}) for multimodal reasoning, including sequentially combining both stages for enhanced reasoning.

\begin{figure*}[t]
\begin{center}
\includegraphics[width=0.98\textwidth]{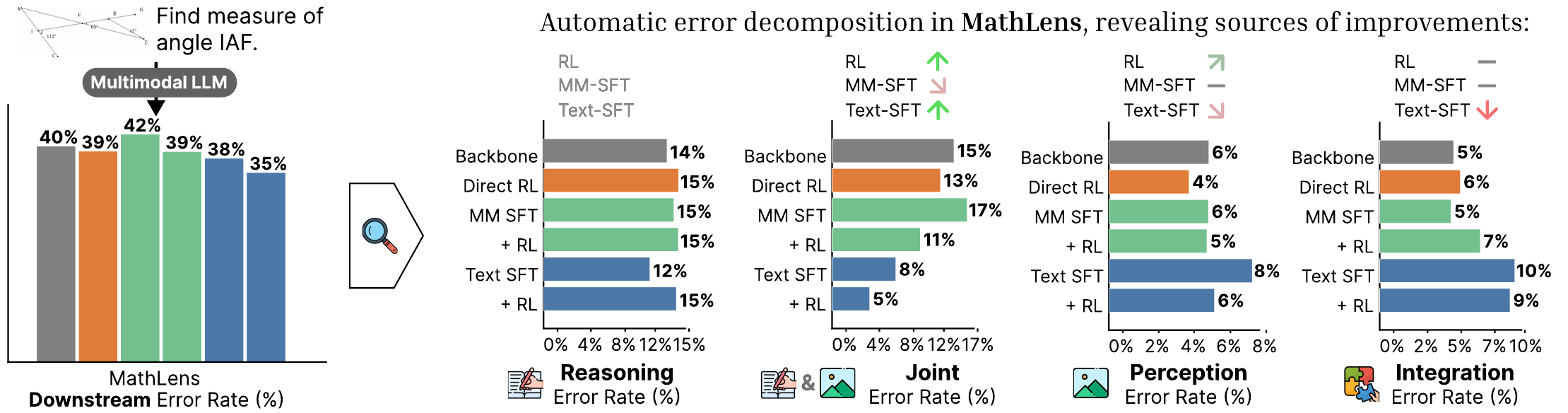}

\caption{\DATANAME decomposes multimodal reasoning errors into perception, reasoning, and a residual multimodal-specific category, revealing component-specific shifts after finetuning that are hidden by aggregate accuracy. Each training strategy affects components differently; e.g., text SFT yields a minor gain (\protect{\color{arrowgreen}$\nearrow$}) in reasoning but a drop (\protect{\color{arrowred}$\downarrow$}) on the multimodal-specific residual (model details in~\cref{sec:ax_impl}).}
\label{fig:teaser_v1}
\end{center}
\end{figure*}

However, unlike LLMs where reasoning-oriented training yields consistent gains, multimodal reasoning training exhibits highly variable outcomes. This motivates analysis of how different training strategies influence specific skills. To this end, multimodal reasoning can be analyzed in terms of perception, reasoning, and multimodal-specific behaviors, each corresponding to a principal source of error.
Existing benchmarks, however, rarely adopt such a decomposition and instead primarily report aggregate accuracy, which obscures the distinctions. Some vary input modalities to approximate skill-specific testing, but without strict controls they fail to isolate components and offer limited diagnostic value (see~\Cref{sec:ax_comparison}). Consequently, it remains unclear which training strategies benefit multimodal reasoning and why.

A central challenge in addressing this question is the lack of domains where such a decomposition can be carried out with explicit semantic control. Geometry provides a natural setting for this analysis: its symbolic structure makes it possible to precisely specify perceptual facts, reasoning operators, and their interaction, and it already underpins most existing multimodal reasoning benchmarks~\citep{lu2024mathvista,wang2024mathvision,zhang2024mathverse}. This level of control is essential for analyzing how training strategies differentially affect perception, reasoning, and their interaction. However, existing benchmarks do not fully exploit this structure to provide controlled, component-wise evaluation.

To close this gap, we present \DATANAME, a controlled benchmark of geometry problems that separately probes perception, reasoning, and multimodal-specific behaviors~(\cref{fig:teaser_v1}).
Starting from the symbolic semantic state of~\citep{zhang2023formalgeo}, we build four aligned annotations~(\cref{subsec:data_gen}): (i) diagrams rendered from geometric constraints to primarily test perception, (ii) a definitionally equivalent textual description to test reasoning under trivialized perception, (iii) multimodal questions requiring both modalities, and (iv) fine-grained probes targeting recovery of visual details.
To further test robustness against visual modifications, \DATANAME introduces semantic diagram modifications that alter visual form while preserving task correctness.
By grounding the benchmark in geometry, \DATANAME retains authentic task complexity and enables rigorous comparison to prior benchmarks.

Across the models we study, \DATANAME demonstrates that training strategies shape multimodal reasoning components in distinct ways.
1) \textit{Perception} is primarily boosted by reinforcement learning, with larger gains when strong textual SFT has already established reasoning competence~(\cref{subsubsec:exp_text}), while our analysis suggests that textual SFT itself, despite lacking visual input, indirectly strengthens perception through reflective reasoning~(\cref{subsubsec:exp_perception}).
2) \textit{Reasoning} improves in tandem with perception under RL, but does not exhibit distinct additional gains beyond those coupled improvements~(\cref{subsubsec:exp_error_type}).
3) \textit{Multimodal-specific} errors form the least-reduced category: RL offers little benefit, and as perception and reasoning improve, errors increasingly fall outside the scope of the measured perception and reasoning components and are therefore categorized as multimodal-specific residuals~(\cref{subsubsec:exp_error_type}).
4) \textit{Robustness} diverges across strategies, with RL enhancing consistency under diagram variation, whereas multimodal SFT reduces robustness, likely through overfitting~(\cref{subsubsec:exp_robustness}).

Our contributions:
\begin{itemize}[leftmargin=*,nosep]
    \item \textbf{Framework for multi-axial evaluation of multimodal reasoning:} A new benchmark that separately probes perception, reasoning, and multimodal-specific residuals, enabling analysis beyond aggregate accuracy.
    \item \textbf{Findings from controlled analysis:} Discoveries about how different training objectives and data settings influence the capabilities of Multimodal Language Models.
    \item \textbf{\DATANAME dataset:} A curated benchmark with problems, annotations, and perturbations.
\end{itemize}

\begin{figure*}[t]
\begin{center}
\includegraphics[width=0.98\textwidth]{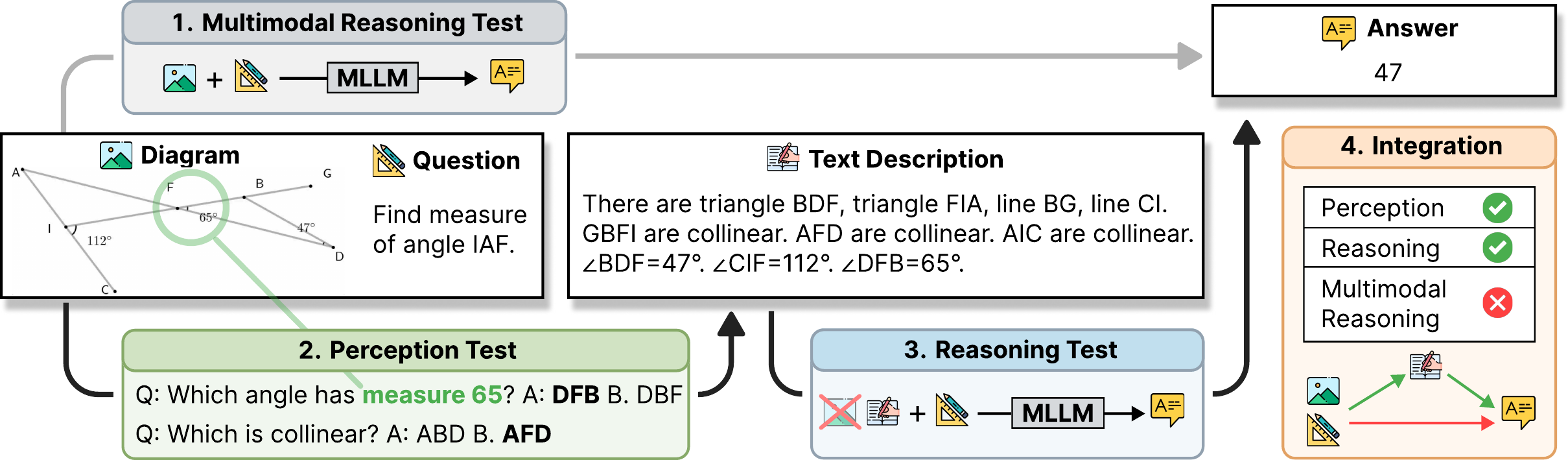}
\end{center}
\caption{Based on \DATANAME annotations, the joint \textit{Multimodal Reasoning Test} (1) is decomposed into a Perception Test and a textual Reasoning Test. The \textit{Perception Test} evaluates questions answerable directly from the diagram, such as reading an annotated angle (e.g., $\angle DFB$ in (2)). The \textit{Reasoning Test} (3) replaces the diagram with a complete textual description (e.g., ``There are triangle $BDF$ \ldots $\angle DFB = 65^\circ$''), such that the question can be solved without visual access. Finally, the \textit{Multimodal-specific} category (4) highlights cases where multimodal reasoning fails even though perception and reasoning, when tested independently, succeed.}
\label{fig:task_v1}
\end{figure*}

\section{\DATANAME}
\label{sec:data}

Most multimodal reasoning benchmarks report only aggregate accuracy, obscuring whether errors stem from perception (extracting information from inputs), reasoning (operating on extracted information), or their interaction. Our benchmark is designed to \textit{separately probe these components} while preserving realistic problem contexts.
\DATANAME comprises 926 geometry problems, each presented with eight diagram variants and an average of $\sim 7.03$ visual probes per problem.

\subsection{Dataset formalization}
\label{subsec:data_form}

\paragraph{Definitions.}
For a problem instance $k$, we consider two latent generative variables: the context semantics $S_k$ and the query operator $\varphi_k$. 
The semantics decomposes atomically,
$S_k=\{s_{k,1},\ldots,s_{k,m}\}$,
where each $s_{k,i}$ encodes a basic fact. For example, in a geometry problem $k$, $s_{k,1}$ could represent (\textsc{$\angle ABC=50^\circ$}).
Both visual diagram $C^{img}_k$ and textual description $C^{txt}_k$ context are generated conditionally on $S_k$,
\[
C^{\mathrm{img}}_k \sim p(C^{\mathrm{img}}\mid S_k),\qquad
C^{\mathrm{txt}}_k \sim p(C^{\mathrm{txt}}\mid S_k).
\]
The surface question $Q_k$ is generated from $\varphi_k$ and a subset of atoms,
\[
Q_k \sim p\bigl(Q \mid \varphi_k,\, S_k^{\text{q}}\bigr),\qquad S_k^{\text{q}}\subseteq S_k,
\]
with $\varphi_k=$ \textsc{(compute $\angle A$)} and $S_k^{\text{q}}=\{s_{k,1}\}$, the question can be ``$\angle ABC=50^\circ$, what is $\angle A$?''
The ground-truth answer is defined by applying the operator to the full semantic state $A_k=f(\varphi_k,S_k)$.

Further, we generate a perception probe set $Q^{\mathrm{perc}}_k$ with semantic atoms $s_{k,i}$. Let
\[
Q^{\mathrm{perc}}_k=\{q^{\mathrm{perc}}_{k,i}\}_{i=1}^m,\qquad
a^{\mathrm{perc}}_{k,i}=\llbracket s_{k,i}\rrbracket,
\]
so each probe targets a single atom of $S_k$ and its gold answer is the atom's valuation.
For example, $q^{\mathrm{perc}}_{k,1} =$ ``Which angle has measure of 50? A. ABC B. ABD'' with $a^{\mathrm{perc}}_{k,1} =$ A.
These probes directly test whether a model has recovered the components of $S_k$ from the observed context.

\paragraph{Isolating components with $S_k$ access.}
Having access to the semantic state $S_k$ allows us to design tests that \textit{separately probe} perception, reasoning, and multimodal-specific behaviors (\Cref{fig:task_v1}).

\textit{Perception.} 
Probe questions $Q^{\mathrm{perc}}k$ with gold answers $a^{\mathrm{perc}}{k,i}$ are derived from the atomic facts listed in $S_k$.
These probes check if the model can recover the specific facts in $S_k$ that are needed to solve the problem from the given input.
Errors here indicate perceptual failures.

\textit{Reasoning.} Given $S_k$, we render a textual description $C^{\mathrm{txt}}_k$ that directly encodes the relevant details. 
Evaluating on $(C^{\mathrm{txt}}_k, Q_k)$ trivializes perception, so the task reduces to applying $\varphi_k$ correctly. 
Errors here isolate reasoning competence, free from perceptual confounds.

\textit{Multimodal-specific (residual).}
We isolate multimodal-specific behaviors by combining accuracy on $(C^{\mathrm{img}}_k, Q_k)$ with auxiliary perception and reasoning measures.
Conditional on success in perception probes and text-only reasoning, any remaining errors on the full task are treated as multimodal-specific residuals; this defines the category operationally, as what is left after controlling for the measured perception and reasoning components. We refer to this category as ``integration'' interchangeably for brevity.

Conceptually, this is a compositionality gap~\citep{press2023measuring, redhardt2025scaling}: the model can perform the required perceptual and reasoning operations separately, but fails when it must compose them to solve the original multimodal task. For example, a model may correctly answer every perception probe of a problem and solve the problem from its textual description, yet still answer the original diagram-based question incorrectly (\Cref{fig:task_v1}, (4)). Such a failure cannot be resolved by improving perception or reasoning alone.

Under this definition, the category captures failures not explained by the measured perception probes and text-only reasoning.

\paragraph{Additional uses of $S_k$.}
First, $S_k$ allows construction of questions that \emph{require context}. 
We define $Q'_k$ by restricting its atoms to exclude those in the context latent $S_k^{\text{c}}$:
\[
Q'_k \sim p(Q \mid \varphi_k, S_k^{\text{q}'}), \qquad 
S_k^{\text{q}'} \subseteq S_k, \qquad 
S_k^{\text{q}'} \cap S_k^{\text{c}} = \varnothing.
\]
By design, $Q'_k$ cannot be answered from the query alone and thus forces reliance on $C_k$.

Second, the symbolic representation $S_k$ allows us to apply \emph{semantic perturbations} of the context. 
Let $\tau \in \mathcal{T}_{\mathrm{AP}}$ denote a transformation from the set of admissible perturbations (e.g., relabeling points or rotating a diagram).  
Applying $\tau$ yields a perturbed specification $S'_k = \tau(S_k)$, while preserving the problem semantics so that
\[
f(\varphi_k, S_k) = f(\varphi_k, S'_k).
\]
Re-rendered contexts ($
C^{\mathrm{img}}_{k'} \sim p(C^{\mathrm{img}} \mid S'_k)$
and
$C^{\mathrm{txt}}_{k'} \sim p(C^{\mathrm{txt}} \mid S'_k)$)
serve as systematic distractors. We measure robustness by prediction agreement across contexts from $S_k$ and $S'_k$, testing semantic invariance beyond pixel-level augmentation.
Pixel-level perturbations are insufficient, since they often cause prediction shifts driven by abnormal appearance rather than semantic change.

\begin{figure*}[t]
\begin{center}

\includegraphics[width=0.98\textwidth]{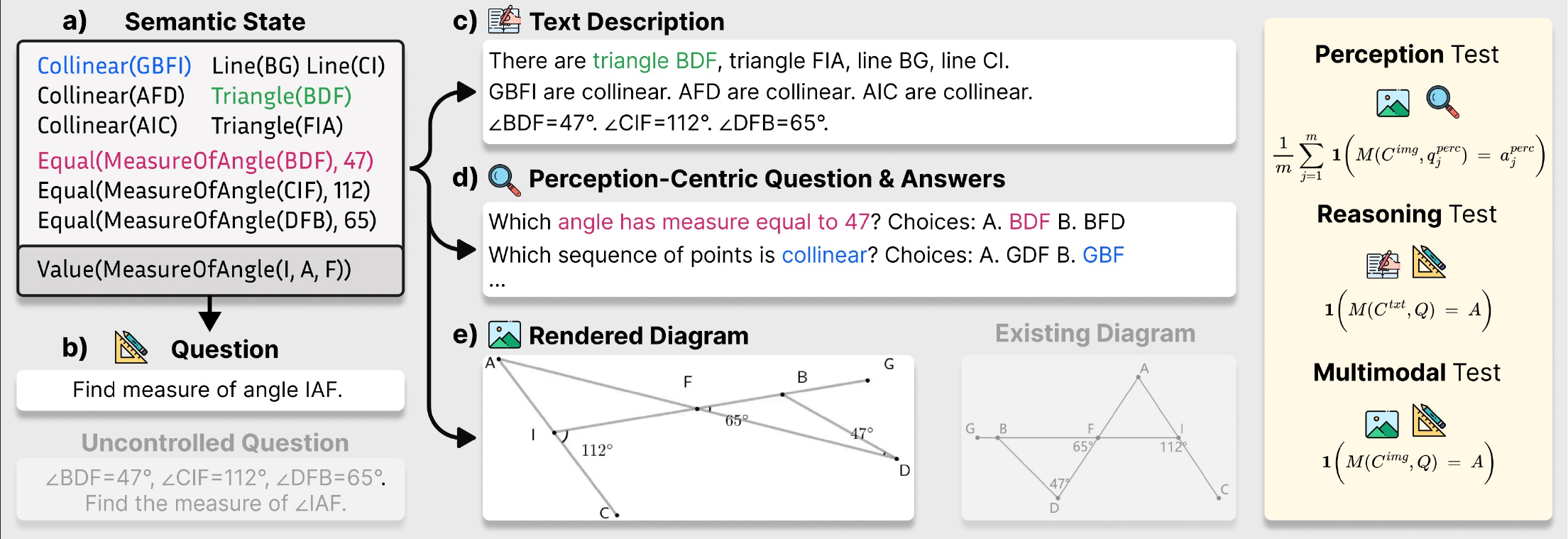}
\end{center}
\caption{\textbf{Sample data generation process in \DATANAME}. From a \textit{semantic state} representation, we build controlled \textit{text descriptions}, \textit{perception probes}, and \textit{questions} with no overlap with visual content. Also, new \textit{diagrams} are rendered from the semantic state to avoid visual familiarity effects\protect\footnotemark.}
\label{fig:data_main}
\end{figure*}
\footnotetext{Here, an obtuse angle is drawn acute to prevent visual estimation and enforce geometric deduction, consistent with Olympiad practice (e.g., ``not drawn to scale'' in AMC).}

\subsection{Data generation pipeline}
\label{subsec:data_gen}

Thus, we build on latent semantics $S_k$ for a set of practical multimodal reasoning problems, matching problem domain and complexity with popular benchmarks in the literature~(\cite{zhang2024mathverse,lu2024mathvista}). Each instance is first specified symbolically as $S_k$ and $\varphi_k$, then rendered into its observable forms: diagrams $C^{\mathrm{img}}_k$, textual descriptions $C^{\mathrm{txt}}_k$, and multiple types of questions.

\paragraph{Data source (\cref{fig:data_main} a).}  
We build on FormalGeo-7K~(\cite{zhang2023formalgeo}), which provides symbolic annotations for geometry problems. 
Each diagram cue or condition is encoded as a predicate (e.g., $\textsc{Collinear}(AB,BC)$, $\textsc{EqualLength}(AB,CD)$), forming the semantic state $S_k$. 
This yields realistic problems with formal representations from which we build the required artifacts.

\paragraph{Questions (\cref{fig:data_main} b).}
We represent each question $Q_k$ as clauses $[S_k^{\text{q}}; \varphi_k]$, with facts $S_k^{\text{q}}$ and goal operator $\varphi_k$. 
For a strictly multimodal question, we drop clauses overlapping with the context $S_k^{\text{c}}$:
\[
Q'_k = f_\text{q}\bigl([(S_k^{\text{q}} \setminus S_k^{\text{c}}); \varphi_k]\bigr),
\]
where $f_\text{q}$ linearizes them into natural language. 
Thus $Q'_k$ requires contextual information to solve.

\paragraph{Textual descriptions (\cref{fig:data_main} c).}
Detail $s_{k,i}\in S_k$ is mapped by template function $f_\text{d}$ to clause,
\[
C^{\mathrm{txt}}_k = \text{concat}_i f_\text{d}(s_{k,i}),
\]
yielding a faithful textual rendering of formal representation $S_k$ without stylistic variation (e.g., $\textsc{EqualLength}(AB,CD)\!\to$ ``Segment $AB$ is equal in length to segment $CD$'').

\paragraph{Perception probes (\cref{fig:data_main} d).}
Each atomic detail $s_{k,i}\!\in\! S_k$ is converted to a probe via templating function $f_\text{p}$ with gold answer $\llbracket s_{k,i}\rrbracket$:  
\[
Q^{\mathrm{perc}}_k = \{f_\text{p}(s_{k,i})\}_i, 
\qquad 
a^{\mathrm{perc}}_{k,i} = \llbracket s_{k,i} \rrbracket.
\]
For example, $\textsc{Collinear}(AB,BC)$ yields ``Which points are collinear? A. ABC B. ABD'' with answer A and negative B.
Thus probes directly test recovery of $S_k$ from context.
The probes cover all diagram information required by the formal solution (\Cref{sec:ax_probes}), so a multimodal-specific error cannot be explained by an omitted atomic fact of $S_k$. The only perceptual difficulties it may include are lower-level ones not represented as symbolic relations, such as localizing the relevant geometric entities in the diagram.

\paragraph{Diagram rendering (\cref{fig:data_main} e).}
Each clause $s_{k,i}\!\in\! S_k$ is converted into geometric constraints (e.g., $\textsc{Perpendicular}(AB,BC)$ $\to (x_a-x_b)(x_c-x_b)+(y_a-y_b)(y_c-y_b)=0$). 
A numerical solver computes coordinates that satisfy all constraints, which are then rendered into diagrams.
We then manually filter outputs to remove artifacts such as occlusions or overlaps.
Details of the manual filtering, including discard rates and annotator agreement, are given in \Cref{sec:ax_annotation}.

\paragraph{Diagram modifications.}
To test visual robustness, we generate diagram variants by altering the semantic state $S_k$, including adding auxiliary elements, applying flips or rotations, merging instances, and relabeling points, while preserving the ground-truth answer. All variants are manually screened.

\subsection{\DATANAMEG: A non-symbolic set}
\label{subsec:data_general}

The main dataset, \DATANAME, focuses on geometry problems, a domain selected intentionally for its well-defined symbolic structure and precise experimental controllability. This choice is methodological: geometry provides a natural and practical setting for isolating and analyzing the multimodal reasoning mechanisms studied in this work.

In addition, we construct \DATANAMEG, a supplementary collection of multimodal reasoning problems drawn from diverse domains. Relative to \DATANAME, this dataset necessarily trades formal semantic rigor for broader domain inclusion, reflecting the intrinsic differences between geometry and less structured multimodal reasoning settings. 
As a result, \DATANAMEG does not admit explicit semantic state specifications; instead, it is curated from prior sources through a consistent and rigorous filtering pipeline to ensure reliability and diversity.
Curation procedures and experimental results are provided in~\Cref{sec:ax_data,sec:ax_exp}.

\begin{figure*}[t]
\begin{center}
\includegraphics[width=0.99\textwidth]{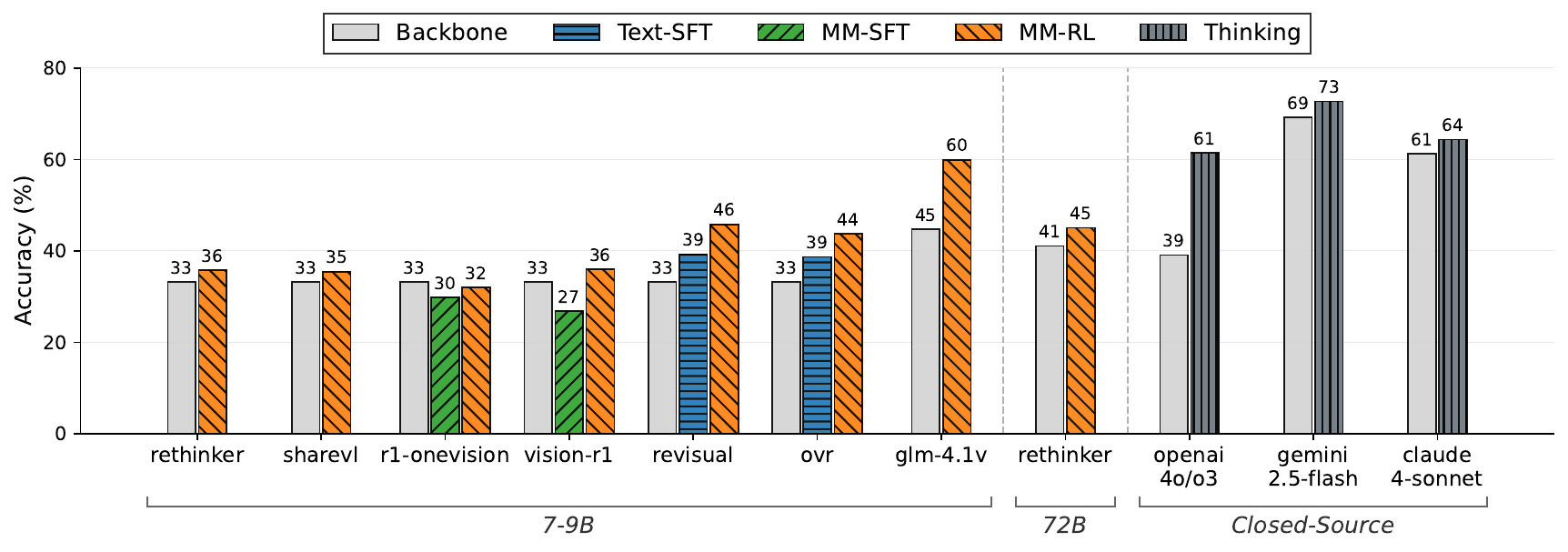}
\end{center}
\caption{\textbf{Impact of multimodal reasoning training on \DATANAME}. We evaluate pretrained backbones alongside models finetuned for multimodal reasoning, reporting accuracy (\%).
\DATANAME is sensitive to gains from multimodal reasoning-oriented finetuning. Inference settings for API models are given in \Cref{sec:ax_impl}.}
\label{fig:main}
\end{figure*}

\section{Experiments}
\label{sec:exp}

We study open-weight multimodal reasoning models in the 7-9B range and their backbone counterparts, motivated by the transparency of open models and the fact that most released multimodal reasoners fall within this scale. We evaluate seven model families, totaling 13 models, including both SFT-only and RL-finetuned variants where available.
We include Qwen-2.5-VL~\citep{bai2025qwen2} and GLM-4.1V-Base~\citep{hong2025glm} as backbone models. Direct multimodal Reinforcement Learning (RL) models comprise VL-Rethinker~\citep{wang2025rethinker} and ShareVL-R1~\citep{yao2025sharevl}. Models trained with multimodal Supervised Finetuning (SFT) followed by multimodal RL include Vision-R1~\citep{huang2025visionr1} and R1-Onevision~\citep{yang2025r1onevision}, while those using textual SFT followed by multimodal RL include Revisual-R1~\citep{chen2025revisual}\footnote{Revisual-R1 adds additional textual RL, yet we group it as textual SFT $\rightarrow$ multimodal RL for consistency.  
} and Open-Vision-Reasoner (OVR)~\citep{wei2025ovr}.
Finally, GLM-4.1V-Thinking~\citep{hong2025glm} falls outside these categories because its training data is undisclosed.

We additionally evaluate six larger models, yielding eight runs in total by including both Gemini-2.5-Flash~\citep{comanici2025gemini} and Claude 4 Sonnet~\citep{anthropic2024claude4} in ``thinking'' and ``non-thinking'' modes: Qwen-2.5-VL-72B, VL-Rethinker-72B, Gemini-2.5-Flash, Claude 4 Sonnet, and GPT-O3/4O~\citep{openai2025introducingo3}.
For each model, we follow the recommended decoding configurations
to generate both reasoning and final answers. Most models employ greedy decoding and yield deterministic outputs. Full experimental details and results are in~\Cref{subsec:ax_exp_full}.

\paragraph{Weak textual reasoning in existing multimodal SFT models.}
We take it as baseline that current multimodal SFT models show limited reasoning following~\cite{chen2025revisual}: their data are easier than text-only sets, leading to weaker performance on reasoning benchmarks. Building stronger data is hindered by the lack of open multimodal reasoning traces (see~\Cref{subsec:ax_prelim} for discussion).

\paragraph{Statistical significance.}
We report paired bootstrap confidence intervals over problem instances for all pre-post finetuning deltas (\cref{fig:text}, \ref{fig:perception}, and \ref{fig:diagram_ood}). Since decoding is deterministic unless stated otherwise, these intervals reflect benchmark sampling variability, not decoding stochasticity. Raw accuracies and diagnostic visualizations (\cref{fig:main}, \ref{fig:robustness}, and \ref{fig:perception_category}) are reported descriptively.

\subsection{Validating \DATANAME}
\label{subsubsec:exp_validity}

Before analyzing finetuning effects, we validate \DATANAME as a multimodal reasoning benchmark by showing it captures the same skills as established benchmarks.
\begin{figure}[t]
  \centering
  
  \begin{minipage}{0.48\textwidth}
    \centering
    \includegraphics[width=\linewidth]{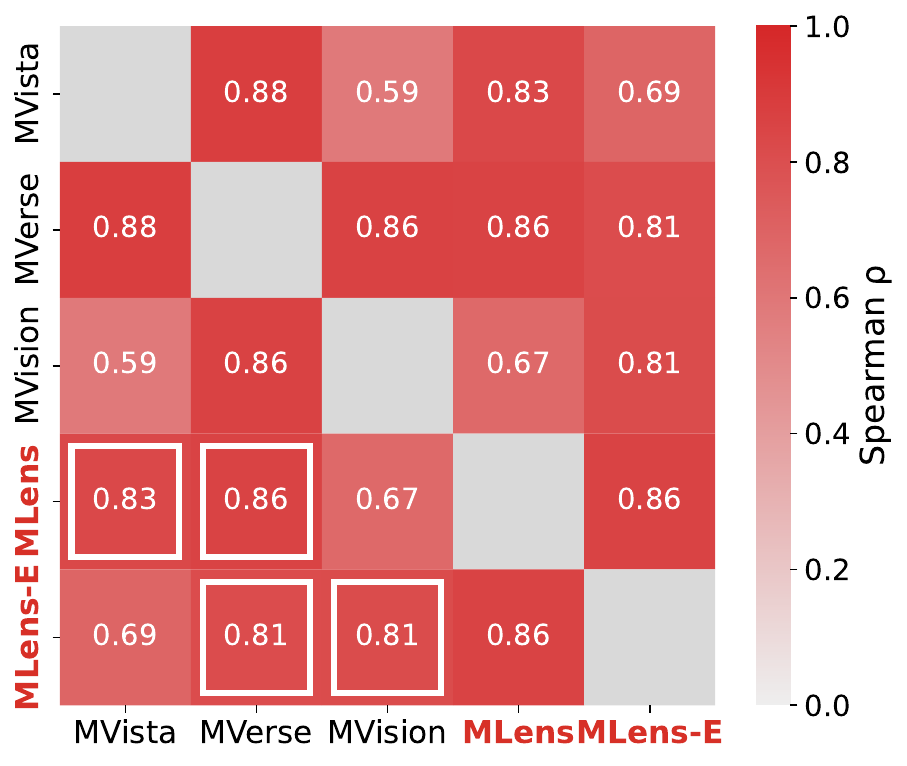}
\caption{\textbf{Correlation of \DATANAME with popular benchmarks.} \DATANAME shows high correlation with standard multimodal reasoning benchmarks as MathVista, MathVerse, and MathVision. We show Spearman correlations between per-model accuracies.}
    \label{fig:corr}
    
  \end{minipage}\hfill %
  \begin{minipage}{0.48\textwidth}
    \centering
\includegraphics[width=\linewidth]{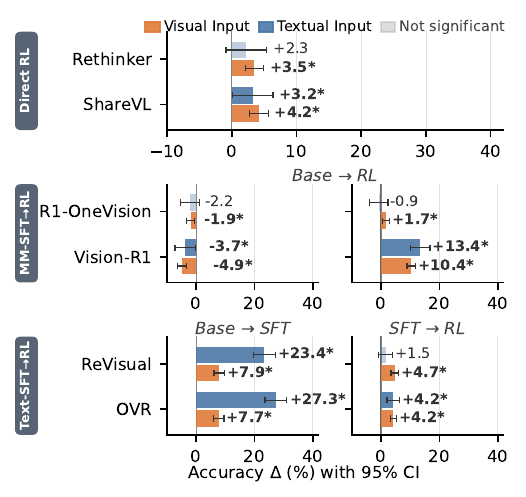}
\vspace{-6mm}
\caption{\textbf{Performance gains by input modality.} Bars show percentage point shifts from finetuning for text versus diagram inputs. Lighter bars are not statistically significant (95\% CI includes zero). Bottom: textual SFT mainly improves performance under textual input, and the subsequent multimodal RL improves both.}
\label{fig:text}
  \end{minipage}
\end{figure}

\paragraph{Sensitivity to finetuning.}
We assess whether \DATANAME reflects gains from multimodal reasoning finetuning by comparing pretrained backbones with their finetuned variants on its downstream task of solving geometry problems from diagrams. As shown in Figure~\ref{fig:main}, finetuned models consistently surpass backbones, demonstrating that \DATANAME is sensitive to these adaptations.

\paragraph{Correlation with established benchmarks.}
To test whether \DATANAME captures patterns consistent with prior benchmarks, we compare it to MathVista~(\cite{lu2024mathvista}), MathVerse~(\cite{zhang2024mathverse}), and MathVision~(\cite{wang2024mathvision}) by correlating model accuracies across eight models (Figure~\ref{fig:corr}, left; full results in Appendix). \DATANAME shows strong Spearman's $\rho$ correlation with MathVista ($\rho=0.83$) and MathVerse ($\rho=0.86$), confirming alignment with established benchmarks, while correlation with MathVision is weaker but still positive ($\rho=0.67$).
Importantly, our goal is not a new downstream benchmark but a controlled, decomposition-focused resource; high correlations therefore underscore its consistency.

\subsection{How much of the gain is explained by improvement in textual reasoning?}
\label{subsubsec:exp_text}

To isolate the role of textual reasoning, we compare performance on textual descriptions $C^{\mathrm{txt}}_k$ and visual diagrams $C^{\mathrm{img}}_k$, which encode identical information conditioned on the question $Q_k$. This parallel annotation lets us track how multimodal training affects each modality. We report accuracy differences (percentage points) before and after multimodal finetuning.

\Cref{fig:text} shows that textual SFT mainly improves textual reasoning, while multimodal RL applied afterward yields larger multimodal gains by shifting its effect to perception and multimodal-specific residuals. Direct multimodal RL without textual SFT gives only modest improvements in both modalities, and poor-quality multimodal SFT degrades multimodal performance more than textual reasoning, with RL only partially recovering the gap.

\begin{tcolorbox}[findingboxstyle]
Finding 1: \textbf{Multimodal RL impact varies with textual reasoning strength.} With strong textual SFT it mainly boosts perception; without it, it modestly improves both modalities (\cref{fig:text}).
\end{tcolorbox}

\subsection{How much of the gain is explained by improvement in perception?}
\label{subsubsec:exp_perception}

\begin{figure}[t]
  \centering
  
  \begin{minipage}{0.48\textwidth}
    \centering
        \includegraphics[width=\linewidth]{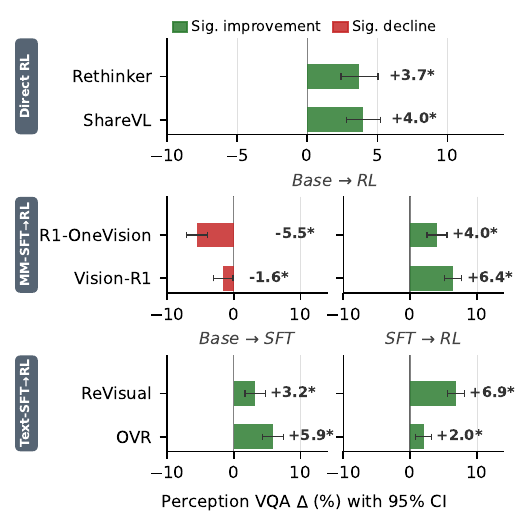}
        \caption{\textbf{Perception performance shifts from finetuning} in percentage points. Except for multimodal SFT, all methods improve perception.} 
        \label{fig:perception}
  \end{minipage}\hfill %
  \begin{minipage}{0.48\textwidth}
    \centering
        \includegraphics[width=\linewidth]{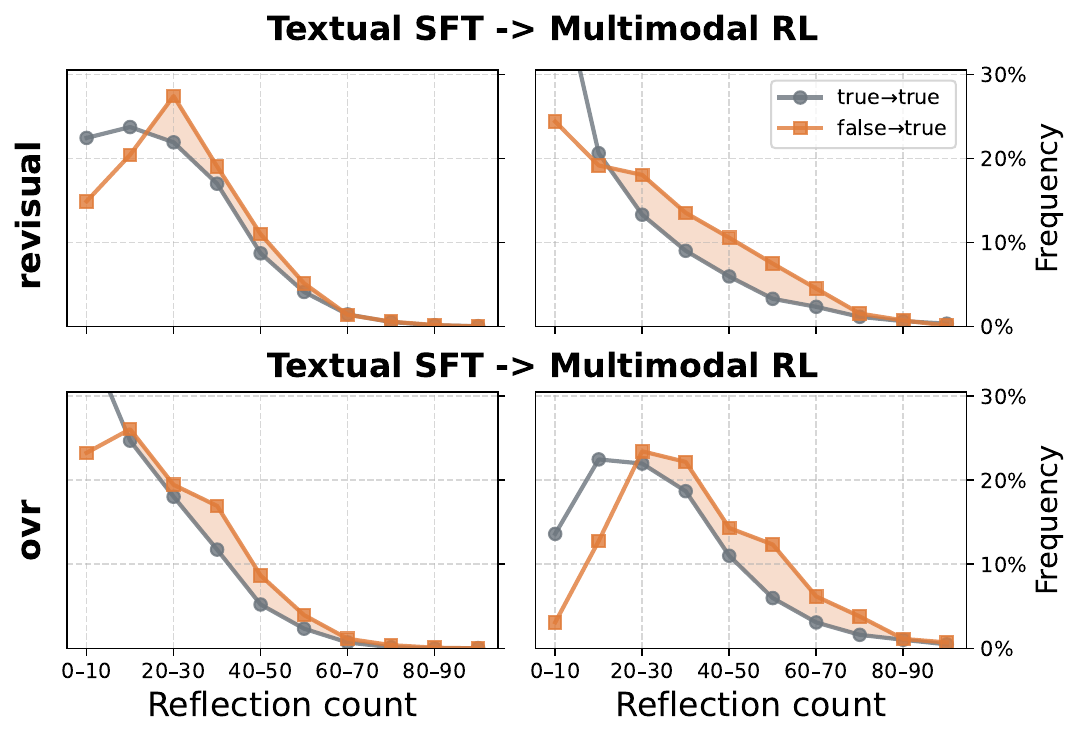}
    \caption{\textbf{Sample distribution over reflection count.} Reflection is more frequent in false→true cases than in true→true cases, showing that perception gains from textual SFT arise partially from reflective reasoning.}
    \label{fig:reflection}
  \end{minipage}
\end{figure}

Next, we evaluate how multimodal reasoning training affects low-level perception. For each problem $Q_k$, we use the perception probes $Q^{\mathrm{perc}}_k = {q^{\mathrm{perc}}_{k,1}, \ldots, q^{\mathrm{perc}}_{k,j}}$.
\Cref{fig:perception} shows that all RL models contribute to better perception required for geometry problem solving. Hence, we conclude that perception is elicited even by training with correctness reward signals on the downstream problems.

\begin{figure*}[t]
\begin{center}
\includegraphics[width=0.98\textwidth, trim=0 0 0.9cm 0, clip]{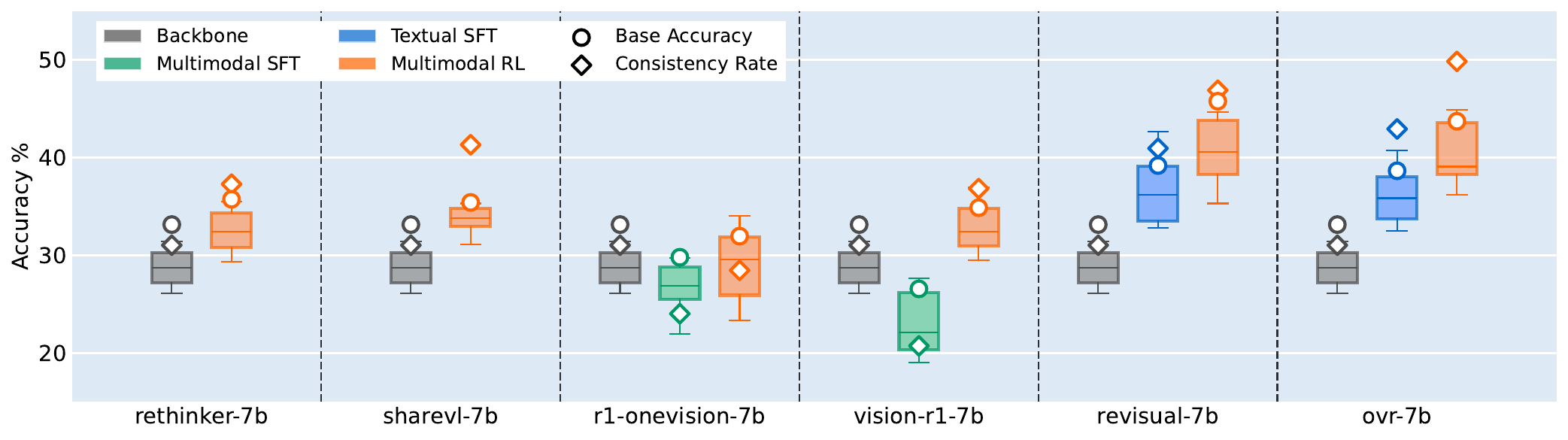}
\end{center}
\caption{\textbf{Robustness to semantic-level visual modifications.} Box plots show accuracy on modified diagrams. Points report accuracy on the unmodified base diagrams and the overall consistency score. Multimodal RL improves consistency, whereas multimodal SFT reduces it.}
\label{fig:robustness}
\end{figure*}

\paragraph{How textual SFT improves perception.}
Textual SFT does not use visual inputs, yet it enhances perceptual performance. These gains suggest influences beyond direct perceptual learning. We hypothesize that one contributing factor is that enhanced reasoning alters how the model interprets ambiguous visual evidence. In particular, stronger reasoning promotes cognitive strategies such as \textit{reflection} and self-correction, which allow the model to revise initial perceptual judgments.  %

We examined the reasoning traces of two models with textual SFT (\Cref{fig:reflection}). Following prior work~\citep{guo2025deepseek, muennighoff2025s1, yang2025speculative}, we measure reflection with a behavioral proxy: the number of ``Wait,'' or ``But wait,'' markers in a response, counting markers at least 30 characters apart. Correct predictions were divided into those accurate both before and after training (true$\rightarrow$true) and those that became accurate only after training (false$\rightarrow$true). We found that reflective reasoning was more frequent in the latter, where accuracy improved post-training.
These results indicate that textual SFT promotes reflective reasoning, which enables models to revisit and correct initial perceptual errors.

\begin{tcolorbox}[findingboxstyle]
Finding 2: \textbf{Textual SFT improves perception} partially through reflective reasoning (\cref{fig:reflection}).
\end{tcolorbox}

\subsection{Error type analysis}
\label{subsubsec:exp_error_type}

\DATANAME's annotation designs enable systematic categorization of model outputs into the following error categories: 
(1) \textit{Perception \& Reasoning}, failures on both perception probes and textual reasoning; (2) \textit{Perception}, failure on perception probes but correct reasoning from text; (3) \textit{Reasoning}, correct perception probes but failed text-based reasoning; (4) \textit{Multimodal-specific}, correct perception and text in isolation but failure on the combined multimodal task.
Correct categories: (5) \textit{Trivial}, solvable from text alone; and (6) \textit{Rest}, all other correct cases.

\Cref{fig:teaser_v1} (see full results in~\Cref{fig:error_types} of Appendix) shows that RL primarily reduces perception and reasoning errors on the same problems, indicating correlated gains. Yet many of the reductions manifest as \textit{Multimodal-specific} residuals, suggesting that finetuning often shifts them into coordination failures rather than eliminating them outright.
\begin{tcolorbox}[findingboxstyle]
Finding 3: \textbf{RL improves perception and reasoning in a correlated manner}, and causes residual errors to be increasingly categorized as multimodal-specific once the other components improve (\cref{fig:teaser_v1}).
\end{tcolorbox}

\subsection{Does multimodal finetuning affect a model's robustness to visual inputs?}
\label{subsubsec:exp_robustness}

\begin{figure}[h]
  \centering
  
  \begin{minipage}{0.48\textwidth}
    We extend the visual familiarity analysis from~\Cref{subsubsec:exp_validity} by testing robustness under controlled diagram variations. 
Instead of pixel-level augmentations (e.g., blurring), we apply \textit{semantic} modifications to the geometric specification, isolating structural variation without introducing low-level artifacts~(see \Cref{sec:ax_data}).  

\paragraph{Downstream accuracy can mask familiarity effects.}
\Cref{fig:diagram_ood} shows that while most training methods are stable under familiar diagrams, multimodal SFT suffers a sharp drop on modified diagrams. This indicates that similar downstream accuracy on public benchmarks can hide reliance on visual familiarity.

\paragraph{RL improves visual consistency.}
To quantify robustness, we use Consistency Rate (CR)~(\cite{zhao2024improving}), the expected agreement of predictions across perturbations of the same diagram:
  \end{minipage}\hfill %
  \begin{minipage}{0.48\textwidth}
    \centering
    \input{figs/inline_shift}
  \end{minipage}
\end{figure}

\begin{align}
\Delta
&=
\mathbf{1}\!\left[
    M(Q'_k, \bar{C}_{k,i}) = M(Q'_k, \bar{C}_{k,j})
\right].\\
CR
&= \mathbb{E}_{Q'_k \sim Q'} \Bigg[
\mathbb{E}_{\substack{\bar{C}_{k,i},\,\bar{C}_{k,j} \sim \bar{C}_k \\ i \neq j}}
\Delta
\Bigg].
\end{align}

where $\bar{C}_k$ is the set of diagram variants for question $Q'_k$.  

\Cref{fig:robustness} shows that accuracy on base diagrams is generally higher than on modified ones, indicating vulnerability to semantic perturbations. Consistency increases after multimodal RL, suggesting robustness to variation. By contrast, multimodal SFT lowers consistency.  %

\begin{tcolorbox}[findingboxstyle]
Finding 4: \textbf{Multimodal RL improves visual consistency} under structural variations (\cref{fig:robustness}).
\end{tcolorbox}

\subsection{Which perception skills benefit from multimodal reasoning training?}
\label{subsubsec:exp_perception_category}

We analyze perception probe performance (\Cref{subsubsec:exp_perception}) by relation type, using a fine-grained categorization of geometric cues (see \Cref{subsec:ax_exp_full} for full results).
This analysis indicates that gains from multimodal reasoning training are not uniform across perceptual skills. Relations such as \textit{cocircular}, \textit{parallel}, and \textit{collinear}, which depend on simple local geometric primitives, show consistent improvement. In contrast, relations such as \textit{same\_angle}, \textit{val\_angle}, \textit{triangle}, and \textit{quadrilateral} benefit only when textual supervision is combined with multimodal reinforcement learning, reflecting their reliance on multi-constraint reasoning and symbol--geometry alignment. By comparison, \textit{same\_length}, \textit{val\_length}, and \textit{perpendicular} remain challenging, likely due to spatially displaced cues or visually ambiguous annotations. Detailed per-category results are reported in the appendix.

\begin{tcolorbox}[findingboxstyle]
Finding 5: \textbf{Perception gains are uneven.} Multimodal reasoning training preferentially improves skills driven by direct geometric cues, while perception tasks that rely on symbolic marks or long-range visual associations show limited gains.
\end{tcolorbox}

\section{\texorpdfstring{Related Work}{Related Work}}
\label{sec:rel_main}

\paragraph{Evaluation of multimodal reasoning.}
Multimodal reasoning benchmarks focus on verifiable domains such as mathematics~\citep{lu2024mathvista,wang2024mathvision,zhang2024mathverse}, scientific diagrams~\citep{yue2024mmmu,hao2025emma}, and charts~\citep{wang2024charxiv}, but most report only a single downstream accuracy per model. Error-type analyses, where provided~\citep{zhang2024mathverse,hao2025emma}, rely on non-standard category definitions and post-hoc inspection of reasoning traces, and the necessity of multimodal context is often left unverified~\citep{yue2024mmmupro}. \DATANAME builds on the formal abstractions of FormalGeo-7K~\citep{zhang2023formalgeo} for fine-grained analysis.

\paragraph{Decomposing capabilities for evaluation.}
Behavioral testing in NLP organizes test cases by the capability they target~\citep{ribeiro2020checklist}, and the compositionality gap measures how often a model answers every sub-question correctly but fails the composed question~\citep{press2023measuring}. For multimodal reasoning, MathVerse~\citep{zhang2024mathverse} and SpatialEval~\citep{wang2024spatialeval} present the same problem with its information shifted between the image and the text; We-Math~\citep{qiao2025wemath} splits problems into sub-problems by knowledge concept; and Prism~\citep{qiao2024prism} separates perception from reasoning by having one model describe the image and another reason over the description. \DATANAME builds aligned diagrams, a definitionally equivalent textual description, and perception probes from the symbolic semantic state of each problem (\Cref{subsec:data_gen}).

\paragraph{Training methods for multimodal reasoning.}
Models are adapted through multimodal SFT, multimodal RL, and textual SFT, often in sequence: multimodal SFT followed by multimodal RL~\citep{huang2025visionr1,yang2025r1onevision}, textual SFT followed by multimodal RL~\citep{chen2025revisual,wei2025ovr}, or direct RL without prior SFT~\citep{deng2025openvlthinker,meng2025mmeureka,wang2025rethinker,yao2025sharevl}. An extended version appears in~\Cref{sec:rel}.

\section{Conclusion}
\label{sec:conclusion}

We introduced \DATANAME, a controlled benchmark that disentangles perception, reasoning, and multimodal-specific behaviors. We find that reinforcement learning primarily improves perception, especially when paired with textual supervision, while textual SFT strengthens perception indirectly through reflective reasoning. Reasoning gains largely follow perceptual improvements, leaving multimodal-specific residuals as a dominant source of remaining error, and robustness diverges across strategies: RL improves consistency under diagram variation, whereas multimodal SFT reduces it due to overfitting.

Looking ahead, our results motivate future work on architectures and training strategies that more directly address cross-modal coordination and perceptual capacity. A discussion of limitations is provided in~\Cref{sec:ax_limitations}.

\section*{Acknowledgements}
We sincerely thank Vidhisha Balachandran, Shivam Garg, Jyoti Aneja, Tyler LaBonte, as well as all members at Microsoft Research AI Frontiers for their invaluable conversations and feedback throughout this project.
This work was conducted in part during an internship at Microsoft Research and in part at Yonsei University. 
The latter was supported by the Institute of Information \& Communications Technology Planning \& Evaluation (IITP) grant funded by the Korean government (MSIT) (No. RS-2020-II201361, Artificial Intelligence Graduate School Program (Yonsei University)).
Additional support was provided by the Institute of Information \& Communications Technology Planning \& Evaluation (IITP) grant funded by the Korean government (MSIT) (No. RS-2021-II211343, Artificial Intelligence Graduate School Program (Seoul National University), No. RS-2026-25522885, Development of a World Foundation Model for Training and Development of Physical AI Systems).
We also acknowledge the KAIT GPU project and the research facilities provided by the ICT at Seoul National University.

{
\small
\bibliographystyle{plainnat}
\bibliography{custom}

\begin{thebibliography}{50}
\providecommand{\natexlab}[1]{#1}
\providecommand{\url}[1]{\texttt{#1}}
\expandafter\ifx\csname urlstyle\endcsname\relax
  \providecommand{\doi}[1]{doi: #1}\else
  \providecommand{\doi}{doi: \begingroup \urlstyle{rm}\Url}\fi

\bibitem[Anthropic(2025)]{anthropic2024claude4}
Anthropic.
\newblock Claude 4 models.
\newblock \url{https://www.anthropic.com/news/claude-4}, 2025.
\newblock Accessed: 2025-09-05.

\bibitem[Austin et~al.(2021)Austin, Odena, Nye, Bosma, Michalewski, Dohan,
  Jiang, Cai, Terry, Le, et~al.]{austin2021program}
Jacob Austin, Augustus Odena, Maxwell Nye, Maarten Bosma, Henryk Michalewski,
  David Dohan, Ellen Jiang, Carrie Cai, Michael Terry, Quoc Le, et~al.
\newblock Program synthesis with large language models.
\newblock \emph{arXiv preprint arXiv:2108.07732}, 2021.

\bibitem[Bai et~al.(2025)Bai, Chen, Liu, Wang, Ge, Song, Dang, Wang, Wang,
  Tang, et~al.]{bai2025qwen2}
Shuai Bai, Keqin Chen, Xuejing Liu, Jialin Wang, Wenbin Ge, Sibo Song, Kai
  Dang, Peng Wang, Shijie Wang, Jun Tang, et~al.
\newblock {Qwen2.5-VL} technical report.
\newblock \emph{arXiv preprint arXiv:2502.13923}, 2025.

\bibitem[Balachandran et~al.(2024)Balachandran, Chen, Joshi, Nushi, Palangi,
  Salinas, Vineet, Woffinden-Luey, and Yousefi]{balachandran2024eureka}
Vidhisha Balachandran, Jingya Chen, Neel Joshi, Besmira Nushi, Hamid Palangi,
  Eduardo Salinas, Vibhav Vineet, James Woffinden-Luey, and Safoora Yousefi.
\newblock {Eureka}: Evaluating and understanding large foundation models.
\newblock \emph{arXiv preprint arXiv:2409.10566}, 2024.

\bibitem[Chen et~al.(2021)Chen, Tworek, Jun, Yuan, Pinto, Kaplan, Edwards,
  Burda, Joseph, Brockman, et~al.]{chen2021evaluating}
Mark Chen, Jerry Tworek, Heewoo Jun, Qiming Yuan, Henrique Ponde De~Oliveira
  Pinto, Jared Kaplan, Harri Edwards, Yuri Burda, Nicholas Joseph, Greg
  Brockman, et~al.
\newblock Evaluating large language models trained on code.
\newblock \emph{arXiv preprint arXiv:2107.03374}, 2021.

\bibitem[Chen et~al.(2025)Chen, Guo, Su, Li, Wu, Chen, Chen, Wang, Qu, and
  Cheng]{chen2025revisual}
Shuang Chen, Yue Guo, Zhaochen Su, Yafu Li, Yulun Wu, Jiacheng Chen, Jiayu
  Chen, Weijie Wang, Xiaoye Qu, and Yu~Cheng.
\newblock Advancing multimodal reasoning: From optimized cold start to staged
  reinforcement learning.
\newblock \emph{arXiv preprint arXiv:2506.04207}, 2025.

\bibitem[Chung et~al.(2025)Chung, Kim, Kim, Lee, Kim, and Yu]{chung2025don}
Jiwan Chung, Junhyeok Kim, Siyeol Kim, Jaeyoung Lee, Min~Soo Kim, and Youngjae
  Yu.
\newblock Don't look only once: Towards multimodal interactive reasoning with
  selective visual revisitation.
\newblock \emph{arXiv preprint arXiv:2505.18842}, 2025.

\bibitem[Comanici et~al.(2025)Comanici, Bieber, Schaekermann, Pasupat,
  Sachdeva, Dhillon, Blistein, Ram, Zhang, Rosen, et~al.]{comanici2025gemini}
Gheorghe Comanici, Eric Bieber, Mike Schaekermann, Ice Pasupat, Noveen
  Sachdeva, Inderjit Dhillon, Marcel Blistein, Ori Ram, Dan Zhang, Evan Rosen,
  et~al.
\newblock {Gemini} 2.5: Pushing the frontier with advanced reasoning,
  multimodality, long context, and next generation agentic capabilities.
\newblock \emph{arXiv preprint arXiv:2507.06261}, 2025.

\bibitem[Dao and Vu(2025)]{dao2025alphamaze}
Alan Dao and Dinh~Bach Vu.
\newblock {AlphaMaze}: Enhancing large language models' spatial intelligence
  via {GRPO}.
\newblock \emph{arXiv preprint arXiv:2502.14669}, 2025.

\bibitem[Deng et~al.(2025)Deng, Bansal, Yin, Peng, Wang, and
  Chang]{deng2025openvlthinker}
Yihe Deng, Hritik Bansal, Fan Yin, Nanyun Peng, Wei Wang, and Kai-Wei Chang.
\newblock {OpenVLThinker}: An early exploration to complex vision-language
  reasoning via iterative self-improvement.
\newblock \emph{arXiv preprint arXiv:2503.17352}, 2025.

\bibitem[Feng et~al.(2025)Feng, Xu, Jiang, Li, Ramasubramanian, Niu, Lin, and
  Poovendran]{feng2025visualsphinx}
Yichen Feng, Zhangchen Xu, Fengqing Jiang, Yuetai Li, Bhaskar Ramasubramanian,
  Luyao Niu, Bill~Yuchen Lin, and Radha Poovendran.
\newblock {VisualSphinx}: Large-scale synthetic vision logic puzzles for {RL}.
\newblock \emph{arXiv preprint arXiv:2505.23977}, 2025.

\bibitem[Fu et~al.(2024)Fu, Hu, Li, Feng, Wang, Lin, Roth, Smith, Ma, and
  Krishna]{fu2024blink}
Xingyu Fu, Yushi Hu, Bangzheng Li, Yu~Feng, Haoyu Wang, Xudong Lin, Dan Roth,
  Noah~A Smith, Wei-Chiu Ma, and Ranjay Krishna.
\newblock {BLINK}: Multimodal large language models can see but not perceive.
\newblock In \emph{ECCV}, pages 148--166. Springer, 2024.

\bibitem[Ghosal et~al.(2024)Ghosal, Han, Ken, and Poria]{ghosal2024language}
Deepanway Ghosal, Vernon Toh~Yan Han, Chia~Yew Ken, and Soujanya Poria.
\newblock Are language models puzzle prodigies? {A}lgorithmic puzzles unveil
  serious challenges in multimodal reasoning.
\newblock \emph{arXiv preprint arXiv:2403.03864}, 2024.

\bibitem[Guo et~al.(2025)Guo, Yang, Zhang, Song, Wang, Zhu, Xu, Zhang, Ma, Bi,
  et~al.]{guo2025deepseek}
Daya Guo, Dejian Yang, Haowei Zhang, Junxiao Song, Peiyi Wang, Qihao Zhu,
  Runxin Xu, Ruoyu Zhang, Shirong Ma, Xiao Bi, et~al.
\newblock {DeepSeek-R1} incentivizes reasoning in {LLMs} through reinforcement
  learning.
\newblock \emph{Nature}, 645\penalty0 (8081):\penalty0 633--638, 2025.

\bibitem[Hao et~al.(2025)Hao, Gu, Wang, Li, Yang, Wang, and Cheng]{hao2025emma}
Yunzhuo Hao, Jiawei Gu, Huichen~Will Wang, Linjie Li, Zhengyuan Yang, Lijuan
  Wang, and Yu~Cheng.
\newblock Can {MLLMs} reason in multimodality? {EMMA}: An enhanced {MultiModal}
  {ReAsoning} benchmark.
\newblock In \emph{ICML}, 2025.

\bibitem[He and {Thinking Machines Lab}(2025)]{he2025nondeterminism}
Horace He and {Thinking Machines Lab}.
\newblock Defeating nondeterminism in llm inference.
\newblock \emph{Thinking Machines Lab: Connectionism}, 2025.
\newblock \doi{10.64434/tml.20250910}.
\newblock
  https://thinkingmachines.ai/blog/defeating-nondeterminism-in-llm-inference/.

\bibitem[He et~al.(2025)He, Liang, Xu, Liu, Chen, Wang, Song, Yu, Liang, Wang,
  et~al.]{he2025deepmath}
Zhiwei He, Tian Liang, Jiahao Xu, Qiuzhi Liu, Xingyu Chen, Yue Wang, Linfeng
  Song, Dian Yu, Zhenwen Liang, Wenxuan Wang, et~al.
\newblock {DeepMath-103K}: A large-scale, challenging, decontaminated, and
  verifiable mathematical dataset for advancing reasoning.
\newblock \emph{arXiv preprint arXiv:2504.11456}, 2025.

\bibitem[Hong et~al.(2025)Hong, Yu, Gu, Wang, Gan, Tang, Cheng, Qi, Ji, Pan,
  et~al.]{hong2025glm}
Wenyi Hong, Wenmeng Yu, Xiaotao Gu, Guo Wang, Guobing Gan, Haomiao Tang, Jiale
  Cheng, Ji~Qi, Junhui Ji, Lihang Pan, et~al.
\newblock {GLM-4.1V-Thinking}: Towards versatile multimodal reasoning with
  scalable reinforcement learning.
\newblock \emph{arXiv preprint arXiv:2507.01006}, 2025.

\bibitem[Huang et~al.(2025)Huang, Jia, Zhai, Cao, Ye, Zhao, Xu, Hu, and
  Lin]{huang2025visionr1}
Wenxuan Huang, Bohan Jia, Zijie Zhai, Shaosheng Cao, Zheyu Ye, Fei Zhao, Zhe
  Xu, Yao Hu, and Shaohui Lin.
\newblock {Vision-R1}: Incentivizing reasoning capability in multimodal large
  language models.
\newblock \emph{arXiv preprint arXiv:2503.06749}, 2025.

\bibitem[Hunter(2007)]{hunter2007matplotlib}
John~D Hunter.
\newblock {Matplotlib}: A {2D} graphics environment.
\newblock \emph{Computing in science \& engineering}, 9\penalty0 (03):\penalty0
  90--95, 2007.

\bibitem[Kwon et~al.(2023)Kwon, Li, Zhuang, Sheng, Zheng, Yu, Gonzalez, Zhang,
  and Stoica]{kwon2023efficient}
Woosuk Kwon, Zhuohan Li, Siyuan Zhuang, Ying Sheng, Lianmin Zheng, Cody~Hao Yu,
  Joseph Gonzalez, Hao Zhang, and Ion Stoica.
\newblock Efficient memory management for large language model serving with
  {PagedAttention}.
\newblock In \emph{SOSP}, pages 611--626, 2023.

\bibitem[Li et~al.(2025)Li, Liu, Li, Zhang, Xu, Chen, Shi, Jiang, Wang, Wang,
  et~al.]{li2025perception}
Yunxin Li, Zhenyu Liu, Zitao Li, Xuanyu Zhang, Zhenran Xu, Xinyu Chen, Haoyuan
  Shi, Shenyuan Jiang, Xintong Wang, Jifang Wang, et~al.
\newblock Perception, reason, think, and plan: A survey on large multimodal
  reasoning models.
\newblock \emph{arXiv preprint arXiv:2505.04921}, 2025.

\bibitem[Lu et~al.(2021)Lu, Gong, Jiang, Qiu, Huang, Liang, and
  Zhu]{lu2021geometry3k}
Pan Lu, Ran Gong, Shibiao Jiang, Liang Qiu, Siyuan Huang, Xiaodan Liang, and
  Song-chun Zhu.
\newblock Inter-gps: Interpretable geometry problem solving with formal
  language and symbolic reasoning.
\newblock In \emph{Proceedings of the 59th Annual Meeting of the Association
  for Computational Linguistics and the 11th International Joint Conference on
  Natural Language Processing (Volume 1: Long Papers)}, pages 6774--6786, 2021.

\bibitem[Lu et~al.(2024)Lu, Bansal, Xia, Liu, Li, Hajishirzi, Cheng, Chang,
  Galley, and Gao]{lu2024mathvista}
Pan Lu, Hritik Bansal, Tony Xia, Jiacheng Liu, Chunyuan Li, Hannaneh
  Hajishirzi, Hao Cheng, Kai-Wei Chang, Michel Galley, and Jianfeng Gao.
\newblock {MathVista}: Evaluating mathematical reasoning of foundation models
  in visual contexts.
\newblock In \emph{ICLR}, 2024.

\bibitem[{Mathematical Association of America}(2025)]{aime2025}
{Mathematical Association of America}.
\newblock Aime i and ii 2025: American invitational mathematics examination.
\newblock
  \url{https://artofproblemsolving.com/wiki/index.php/2025_AIME_I_Problems},
  2025.
\newblock Accessed: 2025-09-05.

\bibitem[Meng et~al.(2025)Meng, Du, Liu, Zhou, Lu, Fu, Han, Shi, Wang, He,
  et~al.]{meng2025mmeureka}
Fanqing Meng, Lingxiao Du, Zongkai Liu, Zhixiang Zhou, Quanfeng Lu, Daocheng
  Fu, Tiancheng Han, Botian Shi, Wenhai Wang, Junjun He, et~al.
\newblock {MM-Eureka}: Exploring the frontiers of multimodal reasoning with
  rule-based reinforcement learning.
\newblock \emph{arXiv preprint arXiv:2503.07365}, 2025.

\bibitem[Muennighoff et~al.(2025)Muennighoff, Yang, Shi, Li, Fei-Fei,
  Hajishirzi, Zettlemoyer, Liang, Cand{\`e}s, and Hashimoto]{muennighoff2025s1}
Niklas Muennighoff, Zitong Yang, Weijia Shi, Xiang~Lisa Li, Li~Fei-Fei,
  Hannaneh Hajishirzi, Luke Zettlemoyer, Percy Liang, Emmanuel Cand{\`e}s, and
  Tatsunori Hashimoto.
\newblock s1: Simple test-time scaling.
\newblock In \emph{Proceedings of the 2025 Conference on Empirical Methods in
  Natural Language Processing}, 2025.

\bibitem[OpenAI(2025)]{openai2025introducingo3}
OpenAI.
\newblock Introducing openai o3 and o4-mini.
\newblock \url{https://openai.com/index/introducing-o3-and-o4-mini/}, 2025.
\newblock Accessed: 2025-09-05.

\bibitem[Press et~al.(2023)Press, Zhang, Min, Schmidt, Smith, and
  Lewis]{press2023measuring}
Ofir Press, Muru Zhang, Sewon Min, Ludwig Schmidt, Noah~A. Smith, and Mike
  Lewis.
\newblock Measuring and narrowing the compositionality gap in language models.
\newblock In \emph{Findings of the Association for Computational Linguistics:
  EMNLP 2023}, 2023.

\bibitem[Qiao et~al.(2025)Qiao, Tan, Dong, Wu, Sun, Song, Wang, GongQue, Lei,
  Zhang, et~al.]{qiao2025wemath}
Runqi Qiao, Qiuna Tan, Guanting Dong, Minhui Wu, Chong Sun, Xiaoshuai Song,
  Jiapeng Wang, Zhuoma GongQue, Shanglin Lei, YiFan Zhang, et~al.
\newblock {We-Math}: Does your large multimodal model achieve human-like
  mathematical reasoning?
\newblock In \emph{Proceedings of the 63rd Annual Meeting of the Association
  for Computational Linguistics}, 2025.

\bibitem[Qiao et~al.(2024)Qiao, Duan, Fang, Yang, Chen, Zhang, Wang, Lin, and
  Chen]{qiao2024prism}
Yuxuan Qiao, Haodong Duan, Xinyu Fang, Junming Yang, Lin Chen, Songyang Zhang,
  Jiaqi Wang, Dahua Lin, and Kai Chen.
\newblock Prism: A framework for decoupling and assessing the capabilities of
  {VLMs}.
\newblock In \emph{Advances in Neural Information Processing Systems}, 2024.

\bibitem[Redhardt et~al.(2025)Redhardt, Akram, and Schug]{redhardt2025scaling}
Florian Redhardt, Yassir Akram, and Simon Schug.
\newblock Scaling can lead to compositional generalization.
\newblock In \emph{Advances in Neural Information Processing Systems}, 2025.

\bibitem[Rein et~al.(2024)Rein, Hou, Stickland, Petty, Pang, Dirani, Michael,
  and Bowman]{rein2024gpqa}
David Rein, Betty~Li Hou, Asa~Cooper Stickland, Jackson Petty, Richard~Yuanzhe
  Pang, Julien Dirani, Julian Michael, and Samuel~R Bowman.
\newblock {GPQA}: A graduate-level {Google}-proof {Q\&A} benchmark.
\newblock In \emph{COLM}, 2024.

\bibitem[Ribeiro et~al.(2020)Ribeiro, Wu, Guestrin, and
  Singh]{ribeiro2020checklist}
Marco~Tulio Ribeiro, Tongshuang Wu, Carlos Guestrin, and Sameer Singh.
\newblock Beyond accuracy: Behavioral testing of {NLP} models with {CheckList}.
\newblock In \emph{Proceedings of the 58th Annual Meeting of the Association
  for Computational Linguistics}, 2020.

\bibitem[Sun et~al.(2025)Sun, Sun, Peng, and Ye]{sun2025mitigating}
Hai-Long Sun, Zhun Sun, Houwen Peng, and Han-Jia Ye.
\newblock Mitigating visual forgetting via take-along visual conditioning for
  multi-modal long {CoT} reasoning.
\newblock \emph{arXiv preprint arXiv:2503.13360}, 2025.

\bibitem[Virtanen et~al.(2020)Virtanen, Gommers, Oliphant, Haberland, Reddy,
  Cournapeau, Burovski, Peterson, Weckesser, Bright, et~al.]{virtanen2020scipy}
Pauli Virtanen, Ralf Gommers, Travis~E Oliphant, Matt Haberland, Tyler Reddy,
  David Cournapeau, Evgeni Burovski, Pearu Peterson, Warren Weckesser, Jonathan
  Bright, et~al.
\newblock {SciPy} 1.0: fundamental algorithms for scientific computing in
  {Python}.
\newblock \emph{Nature methods}, 17\penalty0 (3):\penalty0 261--272, 2020.

\bibitem[Wang et~al.(2025)Wang, Qu, Huang, Chu, Lin, and
  Chen]{wang2025rethinker}
Haozhe Wang, Chao Qu, Zuming Huang, Wei Chu, Fangzhen Lin, and Wenhu Chen.
\newblock {VL-Rethinker}: Incentivizing self-reflection of vision-language
  models with reinforcement learning.
\newblock \emph{arXiv preprint arXiv:2504.08837}, 2025.

\bibitem[Wang et~al.(2024{\natexlab{a}})Wang, Ming, Shi, Vineet, Wang, Li, and
  Joshi]{wang2024spatialeval}
Jiayu Wang, Yifei Ming, Zhenmei Shi, Vibhav Vineet, Xin Wang, Sharon Li, and
  Neel Joshi.
\newblock Is a picture worth a thousand words? {D}elving into spatial reasoning
  for vision language models.
\newblock \emph{NeurIPS}, 37:\penalty0 75392--75421, 2024{\natexlab{a}}.

\bibitem[Wang et~al.(2024{\natexlab{b}})Wang, Pan, Shi, Lu, Ren, Zhou, Zhan,
  and Li]{wang2024mathvision}
Ke~Wang, Junting Pan, Weikang Shi, Zimu Lu, Houxing Ren, Aojun Zhou, Mingjie
  Zhan, and Hongsheng Li.
\newblock Measuring multimodal mathematical reasoning with {MATH-Vision}
  dataset.
\newblock \emph{NeurIPS}, 37:\penalty0 95095--95169, 2024{\natexlab{b}}.

\bibitem[Wang et~al.(2024{\natexlab{c}})Wang, Xia, He, Chen, Liu, Zhu, Liang,
  Wu, Liu, Malladi, et~al.]{wang2024charxiv}
Zirui Wang, Mengzhou Xia, Luxi He, Howard Chen, Yitao Liu, Richard Zhu, Kaiqu
  Liang, Xindi Wu, Haotian Liu, Sadhika Malladi, et~al.
\newblock {CharXiv}: Charting gaps in realistic chart understanding in
  multimodal {LLMs}.
\newblock \emph{NeurIPS}, 37:\penalty0 113569--113697, 2024{\natexlab{c}}.

\bibitem[Wei et~al.(2025)Wei, Zhao, Sun, Lin, Yin, Hu, Zhang, Yu, Lv, Weng,
  et~al.]{wei2025ovr}
Yana Wei, Liang Zhao, Jianjian Sun, Kangheng Lin, Jisheng Yin, Jingcheng Hu,
  Yinmin Zhang, En~Yu, Haoran Lv, Zejia Weng, et~al.
\newblock Open vision reasoner: Transferring linguistic cognitive behavior for
  visual reasoning.
\newblock \emph{arXiv preprint arXiv:2507.05255}, 2025.

\bibitem[Wu and Xie(2024)]{wu2024vstar}
Penghao Wu and Saining Xie.
\newblock {V*}: Guided visual search as a core mechanism in multimodal {LLMs}.
\newblock In \emph{CVPR}, pages 13084--13094, 2024.

\bibitem[Yang et~al.(2025{\natexlab{a}})Yang, Yue, Chaudhary, and
  Han]{yang2025speculative}
Van Yang, Xiang Yue, Vipin Chaudhary, and Xiaotian Han.
\newblock Speculative thinking: Enhancing small-model reasoning with large
  model guidance at inference time.
\newblock In \emph{Conference on Language Modeling (COLM)}, 2025{\natexlab{a}}.

\bibitem[Yang et~al.(2025{\natexlab{b}})Yang, He, Pan, Jiang, Deng, Yang, Lu,
  Yin, Rao, Zhu, et~al.]{yang2025r1onevision}
Yi~Yang, Xiaoxuan He, Hongkun Pan, Xiyan Jiang, Yan Deng, Xingtao Yang, Haoyu
  Lu, Dacheng Yin, Fengyun Rao, Minfeng Zhu, et~al.
\newblock {R1-OneVision}: Advancing generalized multimodal reasoning through
  cross-modal formalization.
\newblock \emph{arXiv preprint arXiv:2503.10615}, 2025{\natexlab{b}}.

\bibitem[Yao et~al.(2025)Yao, Yin, Zhang, Yang, Wang, Wu, Su, Shen, Qiu, Tao,
  et~al.]{yao2025sharevl}
Huanjin Yao, Qixiang Yin, Jingyi Zhang, Min Yang, Yibo Wang, Wenhao Wu, Fei Su,
  Li~Shen, Minghui Qiu, Dacheng Tao, et~al.
\newblock {R1-ShareVL}: Incentivizing reasoning capability of multimodal large
  language models via {Share-GRPO}.
\newblock \emph{arXiv preprint arXiv:2505.16673}, 2025.

\bibitem[Yue et~al.(2024)Yue, Ni, Zhang, Zheng, Liu, Zhang, Stevens, Jiang,
  Ren, Sun, et~al.]{yue2024mmmu}
Xiang Yue, Yuansheng Ni, Kai Zhang, Tianyu Zheng, Ruoqi Liu, Ge~Zhang, Samuel
  Stevens, Dongfu Jiang, Weiming Ren, Yuxuan Sun, et~al.
\newblock {MMMU}: A massive multi-discipline multimodal understanding and
  reasoning benchmark for expert {AGI}.
\newblock In \emph{CVPR}, pages 9556--9567, 2024.

\bibitem[Yue et~al.(2025)Yue, Zheng, Ni, Wang, Zhang, Tong, Sun, Yu, Zhang,
  Sun, et~al.]{yue2024mmmupro}
Xiang Yue, Tianyu Zheng, Yuansheng Ni, Yubo Wang, Kai Zhang, Shengbang Tong,
  Yuxuan Sun, Botao Yu, Ge~Zhang, Huan Sun, et~al.
\newblock {MMMU-Pro}: A more robust multi-discipline multimodal understanding
  benchmark.
\newblock In \emph{Proceedings of the 63rd Annual Meeting of the Association
  for Computational Linguistics}, 2025.

\bibitem[Zhang et~al.(2024)Zhang, Jiang, Zhang, Lin, Guo, Qiu, Zhou, Lu, Chang,
  Qiao, et~al.]{zhang2024mathverse}
Renrui Zhang, Dongzhi Jiang, Yichi Zhang, Haokun Lin, Ziyu Guo, Pengshuo Qiu,
  Aojun Zhou, Pan Lu, Kai-Wei Chang, Yu~Qiao, et~al.
\newblock {MathVerse}: Does your multi-modal {LLM} truly see the diagrams in
  visual math problems?
\newblock In \emph{ECCV}, pages 169--186. Springer, 2024.

\bibitem[Zhang et~al.(2023)Zhang, Zhu, He, Zou, Huang, Jin, Guo, Mao, Li, Zhu,
  et~al.]{zhang2023formalgeo}
Xiaokai Zhang, Na~Zhu, Yiming He, Jia Zou, Qike Huang, Xiaoxiao Jin, Yanjun
  Guo, Chenyang Mao, Yang Li, Zhe Zhu, et~al.
\newblock {FormalGeo}: An extensible formalized framework for olympiad
  geometric problem solving.
\newblock \emph{arXiv preprint arXiv:2310.18021}, 2023.

\bibitem[Zhao et~al.(2024)Zhao, Yan, Sun, Xing, Wang, Meng, Cheng, Ren, and
  Yin]{zhao2024improving}
Yukun Zhao, Lingyong Yan, Weiwei Sun, Guoliang Xing, Shuaiqiang Wang, Chong
  Meng, Zhicong Cheng, Zhaochun Ren, and Dawei Yin.
\newblock Improving the robustness of large language models via consistency
  alignment.
\newblock In \emph{Proceedings of the 2024 Joint International Conference on
  Computational Linguistics, Language Resources and Evaluation (LREC-COLING
  2024)}, pages 8931--8941, 2024.

\end{thebibliography}
}

\newpage

\appendix

\section*{Overview of the Appendix}
\label{sec:ax_overview}

This Appendix is organized as follows:
\begin{itemize}[leftmargin=*]
    \item \Cref{sec:ax_limitations} discusses limitations of this work;
    \item \Cref{sec:rel} contains related work;
    \item \Cref{sec:ax_comparison} contains a comparison with other benchmarks and discusses the benefit of automatic error analysis;
    \item \Cref{sec:ax_data} contains details of the data generation process for the main \DATANAME subset as well as the \DATANAMEG subset;
    \item \Cref{sec:ax_impl} contains implementation details, including hyperparameters and computational resources;
    \item \Cref{sec:ax_exp} contains the complete experimental results corresponding to the main paper figures, as well as results on \DATANAMEG;
    \item \Cref{sec:ax_examples} contains qualitative examples;
    \item \Cref{sec:broader_impact}, \Cref{sec:ethics}, and \Cref{sec:repro} contain the broader impact, ethics, and reproducibility statements.
\end{itemize}

\section{Limitations}
\label{sec:ax_limitations}

\paragraph{Domain.}
\DATANAME is restricted to geometry. We chose geometry because its symbolic structure lets us state every fact in a problem and generate matching diagrams, textual descriptions, and probes, which the decomposition requires. Many other multimodal reasoning tasks, such as charts, scientific figures, and natural images, have no such formal specification. \DATANAMEG covers these domains, but its descriptions and probes are annotated manually, so its decomposition is less controlled.

\paragraph{Public models.}
We compare publicly released models. Models that follow different training strategies also differ in training data, hyperparameters, and other details, so the differences we report cannot be attributed to the training strategy alone. Retraining one backbone under each strategy with matched data would isolate these effects, but requires compute beyond this work.

We encourage future research to extend this framework with broader multimodal domains and controlled retraining protocols.

\section{Related Work}
\label{sec:rel}

\paragraph{Evaluation of multimodal reasoning capacity.}
Multimodal reasoning is a compositional process that requires the integration of perception and abstract reasoning~(\cite{li2025perception}). Inspired by textual reasoning benchmarks, recent multimodal reasoning datasets focus on verifiable domains such as mathematics~(\cite{lu2024mathvista,wang2024mathvision,zhang2024mathverse}), scientific diagrams~(\cite{yue2024mmmu,hao2025emma}), and charts~(\cite{wang2024charxiv}). However, most of these benchmarks report only a single downstream accuracy per model, without systematic means to identify the source of errors. Some further provide manually annotated error-type analyses~(\cite{zhang2024mathverse,hao2025emma}), but these rely on non-standard category definitions across benchmarks, depend on post-hoc semantic inspection of reasoning traces rather than causal diagnosis, and suffer from annotator variance. In addition, the necessity of multimodal context is often left unverified~(\cite{yue2024mmmupro}). Finally, many benchmarks extract problems from publicly available sources such as textbooks, raising risks of data leakage and familiarity bias. To overcome these limitations, our study builds on FormalGeo-7K~(\cite{zhang2023formalgeo}), which provides formal abstractions for both context and goal in practical mathematical geometry problems, enabling rigorous and fine-grained analysis.

\paragraph{Training Methods for Multimodal Reasoning.}
Approaches to adapting MLLMs for multimodal reasoning typically fall into three types: multimodal supervised finetuning, where models are trained on paired image--text inputs with ground-truth reasoning traces and answers; multimodal reinforcement learning, where models are optimized with rewards from verifiable outcomes or reasoning-trace feedback; and textual supervised finetuning, where models are tuned on large-scale textual reasoning corpora without multimodal context. These strategies are often applied sequentially or in combination, such as multimodal SFT followed by multimodal RL (e.g.,~\cite{huang2025visionr1} and~\cite{yang2025r1onevision}) to improve robustness, or textual SFT followed by MM-RL (e.g.,~\cite{chen2025revisual} and ~\cite{wei2025ovr}) to transfer reasoning priors into multimodal domains. In addition, some models adopt direct RL without prior SFT (e.g.,~\cite{deng2025openvlthinker},~\cite{meng2025mmeureka},~\cite{wang2025rethinker}, and~\cite{yao2025sharevl}), demonstrating that reinforcement learning alone can yield competitive reasoning performance. Closed-weight systems (e.g.,~\cite{openai2025introducingo3},~\cite{comanici2025gemini}, and~\cite{anthropic2024claude4}) also report strong multimodal reasoning ability, although their training data and pipelines remain undisclosed.

\section{Comparison with Existing Benchmarks}
\label{sec:ax_comparison}

\begin{figure*}[t]
\begin{center}
\includegraphics[width=0.98\textwidth]{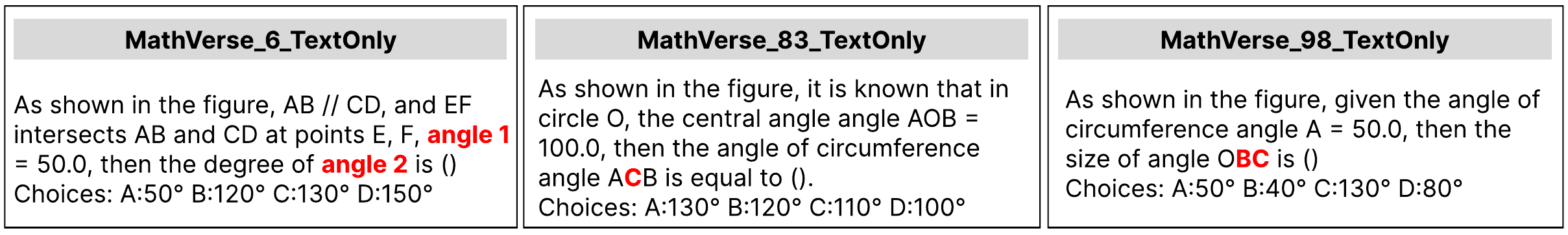}
\end{center}
\caption{\textbf{Example of MathVerse text descriptions.} 
The textual description fails to fully encode geometric relations, requiring external information to be inferred from the diagram. 
This incompleteness makes it unsuitable for evaluating pure reasoning capacity.}
\label{fig:ax_mathverse}
\end{figure*}

\begin{figure*}[t]
\begin{center}
\includegraphics[width=0.98\textwidth]{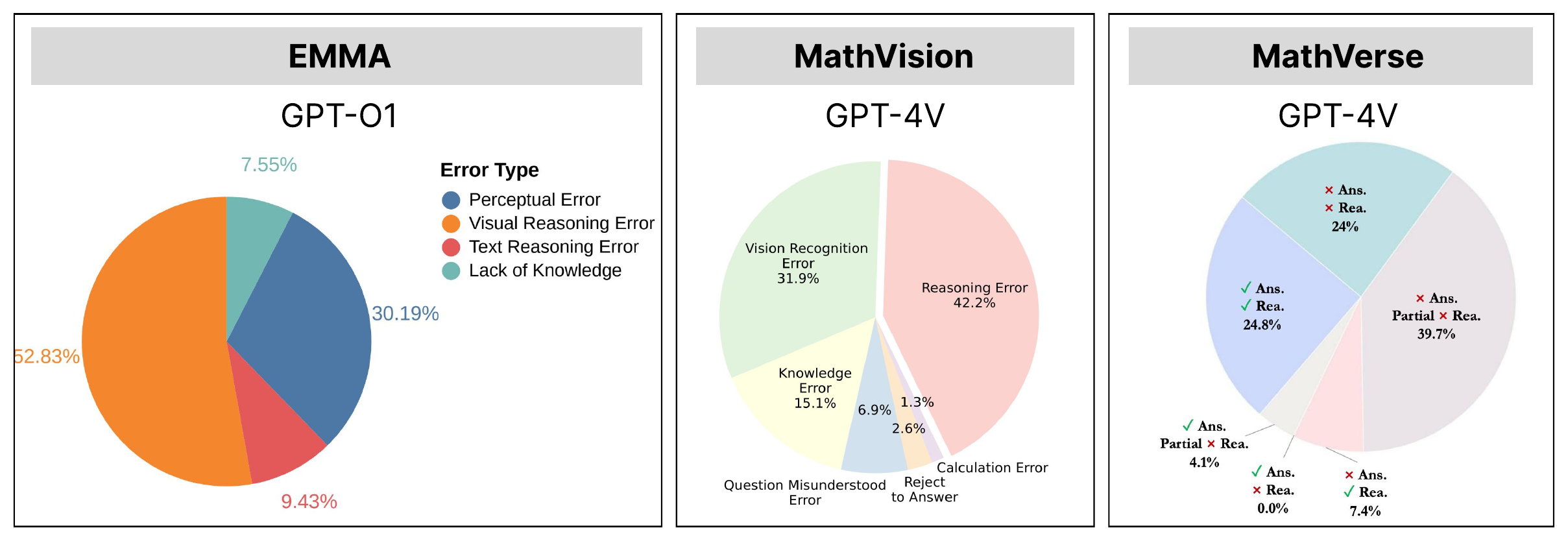}

\caption{\textbf{Inconsistencies in manual error analyses across benchmarks.} Pie charts show variation in error categories across multimodal reasoning datasets, underscoring the lack of standardized criteria. Analyses are typically restricted to a single model. Figures are taken from the respective papers~\citep{hao2025emma,wang2024mathvision,zhang2024mathverse}.}
\label{fig:ax_others_error}
\end{center}
\end{figure*}

\paragraph{Comparison with MathVerse.}
Multimodal mathematical problems are a popular testbed for evaluating multimodal reasoning.  
Among existing benchmarks~(\cite{lu2024mathvista,wang2024mathvision,yue2024mmmu}), the closest to ours is \textit{MathVerse}~(\cite{zhang2024mathverse}), which, like our benchmark, primarily focuses on mathematical geometry problems and partly draws from Geometry3K~(\cite{lu2021geometry3k}), a subset of FormalGeo-7K~(\cite{zhang2023formalgeo}).  
MathVerse also provides multiple input modalities ranging from text-only to vision-only, enabling skill-specific evaluation of multimodal reasoning models, which is conceptually aligned with our design.

However, MathVerse does not meet the criteria necessary for rigorous capacity isolation:  
1) it relies on curated diagrams, making it vulnerable to data leakage or familiarity effects;  
2) its text descriptions are incomplete for evaluating pure reasoning, often requiring external information to be inferred from diagrams (see~\Cref{fig:ax_mathverse});  
3) overlap between question and diagram content is not explicitly controlled.  
Due to these limitations, analyses in this paper cannot be reproduced with the same rigor using prior benchmarks.

\paragraph{Benefits of automatic error analysis.}
Error type analysis offers actionable insights into model weaknesses and guides directions for improvement. However, in multimodal reasoning research, such analyses lack standardization and are almost always performed manually. Consequently, as shown in~\Cref{fig:ax_others_error}, categories and criteria vary across datasets, making it difficult to generalize findings. Moreover, manual analysis is costly and typically applied to a single model, quickly becoming outdated along with its conclusions. By contrast, \DATANAME's automatic analysis pipeline enables consistent, scalable, and up-to-date error categorization across models and datasets.

\section{Data Generation Details}
\label{sec:ax_data}

\subsection{\DATANAME}

\paragraph{Diagram rendering.}
Each clause $s_{k,i}\!\in\! S_k$ is mapped to a corresponding set of algebraic constraints. For instance, $\textsc{Perpendicular}(AB,BC)$ is encoded as ($(x_a-x_b)(x_c-x_b)+(y_a-y_b)(y_c-y_b)=0$). Taken together, the constraints form a nonlinear system defining admissible coordinates for all points. We solve this system using the sequential least-squares programming routine (\texttt{SLSQP}) in \texttt{scipy.optimize.minimize}~(\cite{virtanen2020scipy}), which searches for coordinates that minimize the residual of all constraints subject to feasibility bounds. To improve robustness, initial coordinates are randomly sampled, and optimization is repeated until convergence. Up to ten attempts with different random seeds are allowed, after which the sample is discarded if no feasible solution is found. Valid solutions are rendered with a \texttt{matplotlib}-based backend~(\cite{hunter2007matplotlib}) that draws points, line segments, arcs, and annotations according to the computed geometry. Since automatic optimization may still yield degenerate layouts (e.g., overlapping vertices, occluded labels, or extreme aspect ratios), we apply a post-processing step in which such outputs are manually filtered to preserve clarity and readability.

\paragraph{Diagram modification.}
For each geometry problem, we generate eight diagram variants:  
(1) the \textit{original} diagram from the source dataset,  
(2) a \textit{rendered} diagram generated directly from the symbolic representation,  
and six symbolic modifications:  
(3) \textit{add\_shapes}, which inserts 1--3 random shapes (triangles or quadrilaterals),
(4) \textit{add\_lines}, which inserts 1--3 random lines between existing points,
(5) \textit{flip}, which mirrors the canvas while preserving label orientation,  
(6) \textit{rotate}, which rotates the canvas while keeping labels upright,  
(7) \textit{merge}, which concatenates the current diagram with another randomly chosen one and revises labels accordingly,  
(8) \textit{rename}, which replaces the label set (e.g., A,B,C $\to$ X,Y,Z).
All auto-rendered figures are manually filtered after generation. If any version is invalid or visually unsuitable (e.g., severe occlusions, degenerate angles), the entire problem is discarded.  

\paragraph{Perception probes.}
\label{sec:ax_probes}
Each probe is a binary multiple-choice question, and the order of the two choices is randomized with equal probability. In preliminary experiments with four-way probes, the main conclusions remained qualitatively unchanged, but weaker models showed substantially larger option-position and distractor biases. Consistent with prior observations~\citep{yue2024mmmupro}, these biases became more pronounced as the number of choices increased. We therefore use binary probes throughout.
Probes cover the information required by the formal solution: FormalGeo verifies that its clause set suffices to derive the answer, and every clause is converted into a probe. A multimodal-specific error therefore cannot come from an omitted clause.

\subsection{Manual filtering and annotation quality}
\label{sec:ax_annotation}

\paragraph{Human performance.}
Our curation process discards any problem not solved correctly by the human annotators (validity criterion). This serves as both complexity and quality control, so models are evaluated only on human-solvable problems.

\paragraph{Ambiguity.}
Our annotations are analytically unambiguous, inheriting from the symbolic annotations of FormalGeo. We also discard any sample for which the annotators' answer to a perception probe or to the downstream question diverges from the ground truth.

\paragraph{Annotator agreement.}
Due to the complexity of the multiple criteria, the manual curation process is conducted entirely by two of the authors who understand the requirements in depth. Between them, agreement on the final pass/fail label was 95.8\% (Cohen's $\kappa = 0.89$).

\paragraph{Discard rate.}
\Cref{tab:discard} reports the outcome of the manual screening stage. Candidates are assigned to the first criterion they fail, so the categories are mutually exclusive and the percentages sum to 100\% (up to rounding).

\begin{table}[h]
\centering
\caption{Outcome of manual screening, by the first criterion each candidate fails.}
\label{tab:discard}
\begin{tabular}{lr}
\toprule
Category & Rate (\%) \\
\midrule
Diagram -- Invalid & 23.3 \\
Diagram -- Poor visual & 18.2 \\
Intervention -- Poor visual & 1.3 \\
Question -- Invalid text & 0.5 \\
Question -- Invalid visual & 0.3 \\
Question -- Invalid perception & 0.3 \\
Pass & 56.0 \\
\bottomrule
\end{tabular}
\end{table}

\subsection{\DATANAMEG}
\label{subsec:ax_data_detail}

\DATANAMEG is a curated, re-annotated benchmark that extends the scope of \DATANAME to a broader range of images and problem domains. It includes 107 problems, each paired with an average of $\sim 7.96$ visual probe questions. Representative samples are shown in~\Cref{fig:ax_sample_general1} and~\Cref{fig:ax_sample_general2}, with the data generation procedure detailed below.

\paragraph{Sample collection.}
Problems are sourced from six established multimodal reasoning datasets: BLINK~(\cite{fu2024blink}), V*~(\cite{wu2024vstar}), SpatialEval-Real~(\cite{wang2024spatialeval}), MMMU-Pro~(\cite{yue2024mmmupro}), EMMA~(\cite{hao2025emma}), and MathVista~(\cite{lu2024mathvista}), with mathematical geometry items excluded using metadata.
We retain only multiple-choice problems to maintain consistency. 
Problems requiring more than two images are discarded, and dual-image inputs are concatenated into single images to ensure compatibility with models that do not support multi-image inputs.

\paragraph{Data filtering.}
Problems that appear multimodal may in fact be solvable from text-only correlations~(\cite{yue2024mmmupro}), while others that look complex may reduce to simple pattern matching. 
To guard against such shortcuts, we use model-based validation to test \textit{multimodality} and \textit{reasoning} requirements. 
Each problem is evaluated with Gemini-2.5-Flash~(\cite{comanici2025gemini}) under three settings: (1) full input, (2) text-only input without the image, and (3) full input with reasoning disabled. 
We generate eight responses per setting at temperature~0.6 to capture variability. 

A problem passes the \textit{multimodality} check if text-only accuracy does not exceed chance (1/$k$ for $k$-way choice, e.g., 25\% for 4 options). 
It satisfies the \textit{reasoning} check if accuracy with reasoning disabled falls below chance. 
We enforce \textit{solvability} by discarding problems for which the model fails to answer correctly in all eight full-input attempts. 
Image-level deduplication is then applied to remove visually similar items. 

Finally, human annotators enforce \textit{validity} by independently solving each filtered problem, retaining only those with clearly determinable correct answers. 
This process reduces the initial pool of $\sim$2,000 problems to 200 high-quality samples that meet all four criteria.

\paragraph{Manual annotation.}
We generate seed annotations with Gemini-2.5-Flash, then manually revise them to remove hallucinations and add missing details needed for solvability. 
Textual descriptions are structured in a scene-graph format and decomposed into atomic clauses, from which perception probes are automatically derived. During annotation, we further discard candidates whose required perceptual information cannot be specified exhaustively and unambiguously, yielding the final 107 problems. 

\section{Implementation Details and resources}
\label{sec:ax_impl}

\paragraph{Models.}
Details of the model configurations and corresponding sources are provided in~\Cref{tab:model_hf_groups}.

\begin{table*}[t]
\centering
\begin{tabular}{lll}
\toprule
\textbf{Model} & \textbf{Type} & \textbf{Source} \\
\midrule
\multicolumn{3}{l}{\textbf{7--9B Open-Weight}} \\
\midrule
VL-Rethinker   & backbone  & \texttt{Qwen/Qwen2.5-VL-7B-Instruct} \\
               & mm-rl     & \texttt{TIGER-Lab/VL-Rethinker-7B} \\
ShareVL-R1     & backbone  & \texttt{Qwen/Qwen2.5-VL-7B-Instruct} \\
               & mm-rl     & \texttt{HuanjinYao/R1-ShareVL-7B} \\
R1-OneVision   & backbone  & \texttt{Qwen/Qwen2.5-VL-7B-Instruct} \\
               & mm-sft    & \texttt{Fancy-MLLM/R1-Onevision-7B} \\
               & mm-rl     & \texttt{Fancy-MLLM/R1-Onevision-7B-RL} \\
Vision-R1      & backbone  & \texttt{Qwen/Qwen2.5-VL-7B-Instruct} \\
               & mm-sft    & \texttt{Osilly/Vision-R1-CI-7B} \\
               & mm-rl     & \texttt{Osilly/Vision-R1-7B} \\
Revisual-R1    & backbone  & \texttt{Qwen/Qwen2.5-VL-7B-Instruct} \\
               & text-sft  & \texttt{csfufu/Revisual-R1-Coldstart} \\
               & mm-rl     & \texttt{csfufu/Revisual-R1-final} \\
OVR            & backbone  & \texttt{Qwen/Qwen2.5-VL-7B-Instruct} \\
               & text-sft  & \texttt{Kangheng/OVR-7B-ColdStart} \\
               & mm-rl     & \texttt{Kangheng/OVR-7B-RL} \\
GLM-4.1V       & backbone  & \texttt{zai-org/GLM-4.1V-9B-Base} \\
               & mm-rl     & \texttt{zai-org/GLM-4.1V-9B-Thinking} \\
\midrule
\multicolumn{3}{l}{\textbf{72B Open-Weight}} \\
\midrule
VL-Rethinker      & backbone  & \texttt{Qwen/Qwen2.5-VL-72B-Instruct} \\
               & mm-rl     & \texttt{TIGER-Lab/VL-Rethinker-72B} \\
\midrule
\multicolumn{3}{l}{\textbf{Closed-Weight}} \\
\midrule
OpenAI         & backbone  & GPT-4O (\texttt{gpt-4o\_2024-11-20}) \\
               & thinking  & GPT-O3 (\texttt{o3\_2025-04-16}) \\
Gemini & backbone & \texttt{gemini-2.5-flash (thinking=disabled)} \\
                 & thinking & \texttt{gemini-2.5-flash (thinking=enabled)} \\
Claude & backbone & \texttt{claude-4-sonnet (thinking=disabled)} \\
                 & thinking & \texttt{claude-4-sonnet (thinking=enabled)} \\
\bottomrule
\end{tabular}
\caption{Model configurations studied in this work.}
\label{tab:model_hf_groups}
\end{table*}

\paragraph{Model configuration for~\Cref{fig:teaser_v1}.}
All models are finetuned from Qwen-2.5-VL-7B as the backbone MLLM. VL-Rethinker represents the direct RL setting, Vision-R1 serves as the multimodal SFT model, and Revisual-R1 corresponds to the textual SFT model. Vision-R1 and Revisual-R1 further include their respective RL-extended variants.

\paragraph{Hyperparameters \& computation.}
We use Eureka ML Insights Framework~(\cite{balachandran2024eureka}) for reproducible evaluation.
We run 7--9B parameter models on four NVIDIA A100 80GB GPUs, and 72B models on eight. Generation is accelerated and parallelized with the vLLM~(\cite{kwon2023efficient}) library. Most experiments use greedy decoding (temperature $0.0$) for deterministic outputs. For models that otherwise suffer from text degeneration through severe repetition (e.g.,~\cite{huang2025visionr1}), we apply stochastic decoding with temperature $0.6$ and top-$p$ $0.65$.
The default maximum generation length is 32,768 tokens to accommodate long reasoning chains. For OVR~(\cite{wei2025ovr}), we extend this limit to 48,000 tokens due to frequent truncations at lower cutoffs. API models use provider defaults except as follows: GPT-4o (\texttt{gpt-4o-2024-11-20}) with at most 16,384 output tokens; o3 (\texttt{o3-2025-04-16}) with the default reasoning effort; Gemini-2.5-Flash with at most 16,384 output tokens and a thinking budget of 0 (non-thinking) or 5,000 tokens (thinking); and Claude 4 Sonnet (\texttt{claude-sonnet-4-20250514}) with at most 32,768 output tokens and, in thinking mode, a 30,720-token thinking budget at temperature 1.0.

\paragraph{Visual resources.}
Visual icons used in~\Cref{fig:data_main} are adapted from \url{flaticon.com}.

\paragraph{Large Language Model Usage.}
LLMs (ChatGPT, GPT-4/5 class and Claude 4 Sonnet) were employed to refine phrasing, improve clarity, and standardize style in sections of the manuscript, but all scientific ideas, experiments, and analyses were conceived, executed, and validated by the authors. LLMs were also used in a limited capacity to assist with literature discovery (e.g., surfacing related work for manual screening). All substantive content decisions, experiment design, and result interpretation remain entirely author-driven.

\section{Experiment}
\label{sec:ax_exp}

\subsection{Preliminaries}
\label{subsec:ax_prelim}

\paragraph{Weak textual reasoning in multimodal SFT models.}
Prior work~(\cite{sun2025mitigating}) highlights that multimodal SFT datasets are considerably easier than textual SFT datasets, which contributes to weaker textual reasoning capacity in trained models. 
For example, the Vision-R1 dataset~(\cite{huang2025visionr1}) averages 821.5 tokens per reasoning trace with a 96.0\% pass rate, whereas the text-only DeepMath dataset~(\cite{he2025deepmath}) averages 8,207.8 tokens with a 75.0\% pass rate. 
This difference suggests that multimodal SFT data require substantially less reasoning effort. 
Consistent with this, multimodal SFT models trained on such data underperform on standard reasoning benchmarks compared to textual SFT models. 
Finally, while textual SFT data can be constructed by distilling reasoning traces from large language models~(\cite{guo2025deepseek}), no comparably strong multimodal reasoning models with open reasoning traces currently exist, making effective multimodal SFT data generation particularly challenging.

\subsection{Further Insights}

\paragraph{Insight on visual familiarity effects.} In~\Cref{fig:corr} of the main paper, we additionally evaluate \DATANAME-E, a variant that uses the same geometry problems but replaces \textit{rendered} diagrams with \textit{existing} diagrams from the original sources. 
Interestingly, \DATANAME-E correlates strongly with MathVerse and MathVision ($\rho=0.81$), while showing a weaker correlation with MathVista ($\rho=0.69$).
This difference suggests that models may leverage visual familiarity with diagram styles from textbooks or public tests when tackling MathVision, an advantage that disappears with freshly-rendered diagrams. This underscores that high benchmark accuracy does not necessarily indicate strong multimodal reasoning, as performance may be inflated by visual familiarity effects from training data.

\subsection{Full Results}
\label{subsec:ax_exp_full}

\begin{figure*}[ht]
\begin{center}
\includegraphics[width=0.98\textwidth]{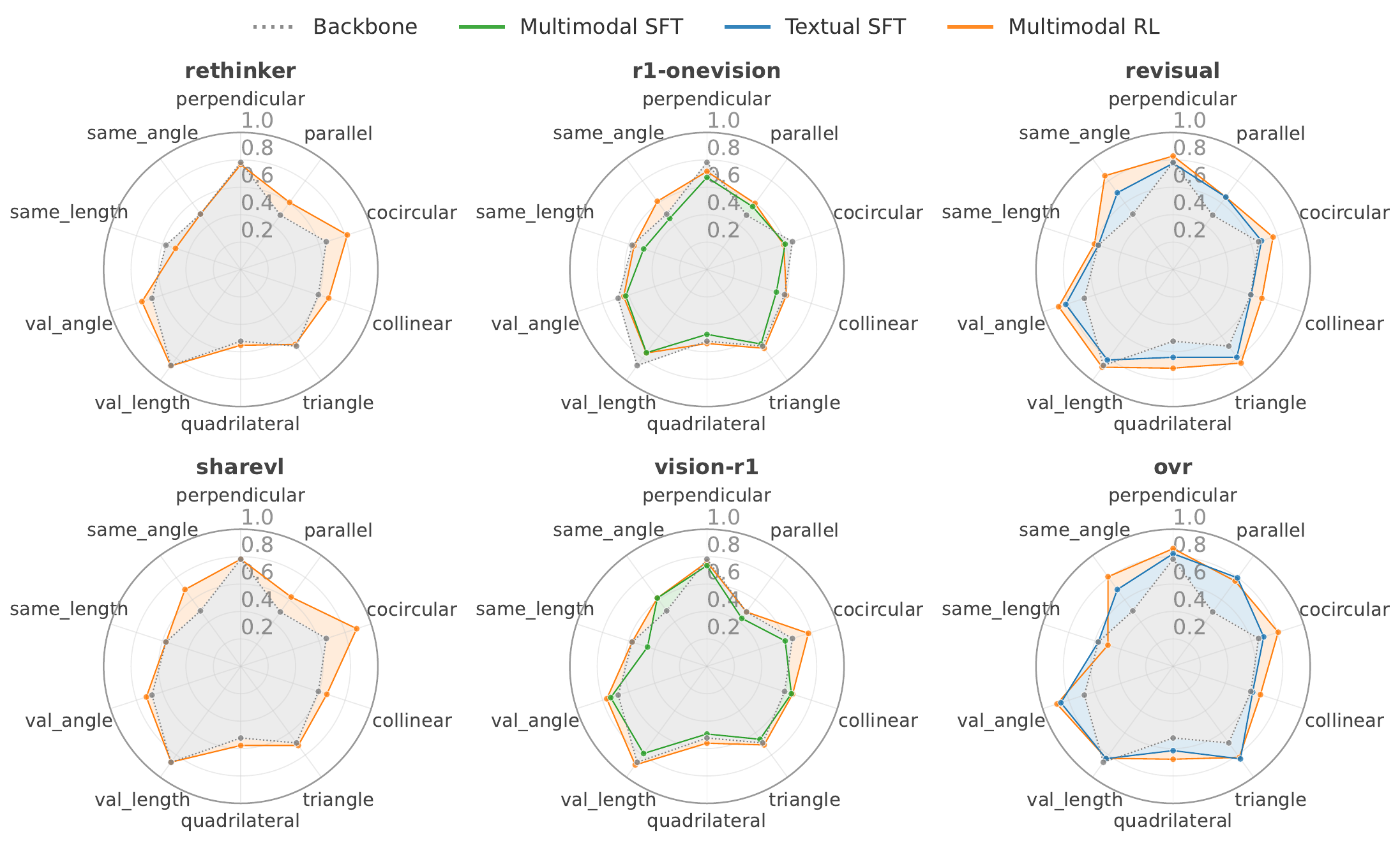}
\end{center}
\caption{\textbf{Perception probe results by question type.} Improvements differ across skills: angle-related and cocircularity tasks improve with multimodal RL, line-length tasks show little change, and polygon detection improves only with textual SFT. A likely factor is whether notations are colocated with the geometric element they describe or spatially separated.}
\label{fig:perception_category}
\end{figure*}

\paragraph{Perception probe results by relation type.}
We report perception probe performance decomposed by question type in~\Cref{fig:perception_category}. Each category groups probes according to the geometric relation being tested (e.g., angle relations, length relations, polygon structure).
Performance gains differ substantially across categories. Angle-related relations and cocircularity show consistent improvement under multimodal reinforcement learning, while line-length relations exhibit limited change across training variants. Polygon-level relations, including triangle and quadrilateral detection, improve primarily when textual supervised finetuning is included.
These differences appear correlated with the visual presentation of the target cue. Relations whose defining annotations are colocated with the relevant geometric element tend to benefit more from multimodal training, whereas relations relying on spatially separated marks or distant comparisons remain challenging.

\begin{table*}
\centering
\begin{tabular}{l|ccc|cc}
\toprule
Model & MathVista & MathVerse & MathVision & \textbf{\DATANAME-E} & \textbf{\DATANAME} \\
\midrule
Qwen2.5-VL-7B & 68.2 & 46.3 & 25.1 & 34.7 & 33.2 \\
R1-Onevision-7B & 64.1 & 46.4 & 29.9 & 35.4 & 29.8 \\
MM-Eureka-Qwen-7B & 73.0 & 50.3 & 26.9 & 34.0 & 31.3 \\
ThinkLite-VL-7B & 74.3 & 52.2 & 29.9 & 34.3 & 32.9 \\
R1-ShareVL-7B & 75.4 & 52.8 & 29.5 & 36.7 & 35.4 \\
VL-Rethinker-7B & 74.9 & 54.2 & 32.3 & 37.6 & 35.7 \\
Qwen2.5-VL-72B & 74.8 & 57.2 & 38.1 & 41.6 & 41.0 \\
VL-Rethinker-72B & 80.4 & 63.5 & 44.9 & 47.6 & 45.0 \\
\bottomrule
\end{tabular}
\caption{Full results of all models used to produce the correlation plot in~\Cref{fig:corr}.}
\label{tab:results_corr}
\end{table*}

\paragraph{Correlation plot details.} 
Table~\ref{tab:results_corr} reports the full benchmark scores for all models used in the correlation analysis of~\Cref{fig:corr}. Results for MathVista, MathVerse, and MathVision are drawn from prior work~(\cite{wang2025rethinker}). Consequently, the set of models differs from our main evaluation and includes additional variants such as MM-Eureka~(\cite{meng2025mmeureka}) and ThinkLite-VL~(\cite{deng2025openvlthinker}). These models were excluded from the main analysis for two reasons: (i) to maintain a balanced number of models across categories, particularly those trained with direct RL, and (ii) to focus on the stronger-performing models on other benchmarks.

\begin{table*}[t]
\centering
\resizebox{\textwidth}{!}{%
\begin{tabular}{l l |c |c |c |c c c c c c |c}
\toprule
\multirow{2}{*}{\textbf{Model}} & \multirow{2}{*}{\textbf{Variant}} & \multirow{2}{*}{Text} & \multirow{2}{*}{Raw} & \multirow{2}{*}{Base} & Add & Add & \multirow{2}{*}{Flip} & \multirow{2}{*}{Merge} & \multirow{2}{*}{Rename} & \multirow{2}{*}{Rotate} & \multirow{2}{*}{Consistency} \\
 &  &  &  &  & lines & shapes &  &  &  &  &  \\
\midrule
\multirow{2}{*}{VL-Rethinker-7B} & Backbone & 38.9 & 34.7 & 33.2 & 26.1 & 28.1 & 30.6 & 26.9 & 29.4 & 31.4 & 31.0 \\
 & MM-RL & 41.1 & 37.6 & 35.7 & 31.4 & 29.4 & 34.7 & 30.6 & 33.5 & 35.5 & 37.3 \\
\midrule
\multirow{2}{*}{ShareVL-R1-7B} & Backbone & 38.9 & 34.7 & 33.2 & 26.1 & 28.1 & 30.6 & 26.9 & 29.4 & 31.4 & 31.0 \\
 & MM-RL & 42.1 & 36.7 & 35.4 & 31.1 & 32.7 & 33.9 & 33.7 & 35.1 & 35.3 & 41.3 \\
\midrule
\multirow{3}{*}{R1-OneVision-7B} & Backbone & 38.9 & 34.7 & 33.2 & 26.1 & 28.1 & 30.6 & 26.9 & 29.4 & 31.4 & 31.0 \\
 & MM-SFT & 36.7 & 35.4 & 29.8 & 21.9 & 26.1 & 29.7 & 25.3 & 27.6 & 29.2 & 24.0 \\
 & MM-RL & 35.9 & 33.0 & 32.0 & 25.1 & 28.5 & 34.0 & 23.3 & 30.7 & 32.3 & 28.5 \\
\midrule
\multirow{3}{*}{Vision-R1-7B} & Backbone & 38.9 & 34.7 & 33.2 & 26.1 & 28.1 & 30.6 & 26.9 & 29.4 & 31.4 & 31.0 \\
 & MM-SFT & 35.2 & 34.3 & 26.8 & 19.1 & 21.7 & 27.8 & 20.6 & 23.1 & 28.0 & 21.1 \\
 & MM-RL & 48.6 & 44.5 & 36.0 & 33.0 & 31.5 & 36.6 & 30.6 & 34.2 & 37.8 & 39.4 \\
\midrule
\multirow{3}{*}{Revisual-R1-7B} & Backbone & 38.9 & 34.7 & 33.2 & 26.1 & 28.1 & 30.6 & 26.9 & 29.4 & 31.4 & 31.0 \\
 & Text-SFT & 62.3 & 43.8 & 39.2 & 32.9 & 35.1 & 42.7 & 32.8 & 37.4 & 39.8 & 41.0 \\
 & MM-RL & 63.8 & 51.7 & 45.8 & 35.3 & 38.0 & 44.4 & 39.1 & 42.2 & 44.7 & 46.9 \\
\midrule
\multirow{3}{*}{OVR-7B} & Backbone & 38.9 & 34.7 & 33.2 & 26.1 & 28.1 & 30.6 & 26.9 & 29.4 & 31.4 & 31.0 \\
 & Text-SFT & 66.2 & 46.7 & 38.7 & 33.8 & 33.7 & 37.9 & 32.5 & 38.1 & 40.7 & 42.9 \\
 & MM-RL & 70.4 & 49.8 & 43.7 & 38.4 & 38.2 & 44.9 & 36.2 & 39.7 & 44.9 & 49.9 \\
\midrule
\multirow{2}{*}{GLM-4.1V-9B} & Backbone & 40.0 & 44.6 & 44.7 & 37.7 & 37.8 & 44.7 & 36.7 & 40.4 & 43.7 & 39.7 \\
 & MM-RL & 69.2 & 65.8 & 59.9 & 52.3 & 50.2 & 60.8 & 50.0 & 57.0 & 61.4 & 62.8 \\
\midrule
\multirow{2}{*}{VL-Rethinker-72B} & Backbone & 52.9 & 41.6 & 41.0 & 36.2 & 35.5 & 42.2 & 37.5 & 38.4 & 41.6 & 40.0 \\
 & MM-RL & 56.2 & 47.6 & 45.0 & 38.2 & 39.1 & 45.5 & 38.4 & 44.6 & 44.8 & 43.9 \\
\midrule
\multirow{2}{*}{OpenAI 4O / O3} & Backbone & 49.1 & 40.7 & 39.1 & 35.2 & 32.9 & 40.7 & 33.6 & 39.0 & 40.8 & 38.7 \\
 & Thinking & 74.5 & 66.8 & 61.4 & 51.9 & 49.9 & 61.7 & 55.8 & 57.8 & 62.9 & 59.0 \\
\midrule
\multirow{2}{*}{Gemini-2.5-Flash} & Backbone & 79.9 & -- & 69.2 & -- & -- & -- & -- & -- & -- & -- \\
 & Thinking & 82.3 & -- & 72.7 & -- & -- & -- & -- & -- & -- & -- \\
\midrule
\multirow{2}{*}{Claude-4-Sonnet} & Backbone & 75.1 & 60.7 & 61.2 & 51.0 & 53.3 & 60.4 & 48.1 & 57.0 & 60.6 & 55.7 \\
 & Thinking & 84.1 & 65.6 & 64.4 & 51.0 & 55.0 & 65.6 & 51.5 & 61.0 & 67.0 & 57.0 \\
\bottomrule
\end{tabular}}
\caption{Downstream accuracy across diagram modifications with output consistency under these conditions. \textit{Text} indicates performance from textual descriptions instead of diagrams, \textit{Raw} denotes original human-generated diagrams, \textit{Base} the newly rendered diagrams, and the other cases semantic-space modifications followed by rendering.}
\label{tab:ax_main_perf}
\end{table*}

\begin{figure*}[t]
\begin{center}
\includegraphics[width=0.9\textwidth]{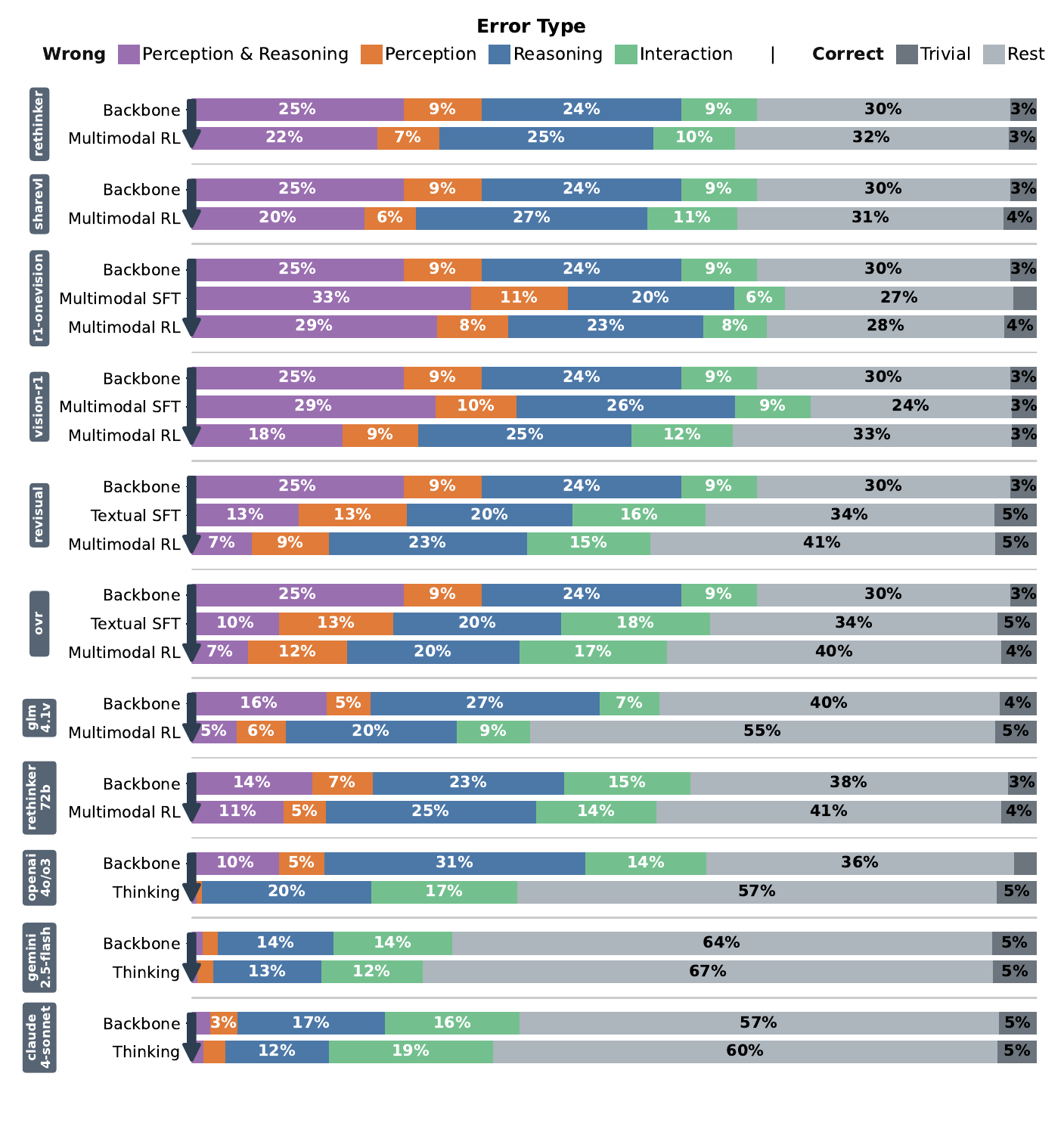}
\end{center}
\vspace{-15pt} %
\caption{\textbf{Distribution of error types per model.} Most gains come from cases failing in both perception and reasoning. Multimodal-specific residual errors become more prominent as the other categories improve.}
\label{fig:error_types}
\end{figure*}

\paragraph{Downstream performance under diagram modifications.}
\Cref{tab:ax_main_perf} reports the complete downstream evaluation results on \DATANAME, including all diagram modifications. These values underlie~\Cref{fig:main},~\Cref{fig:text}, ~\Cref{fig:robustness}, and~\Cref{fig:diagram_ood}. Note that the error-type analysis in~\Cref{fig:error_types} relies on per-sample categorization and cannot be obtained directly from the aggregate scores presented here.

\begin{table*}[t]
\centering
\resizebox{\textwidth}{!}{%
\begin{tabular}{ll| c c c c c c c c c c}
\toprule
\multirow{2}{*}{\textbf{Model}} & \multirow{2}{*}{\textbf{Variant}} & Tri- & Quad- & \multirow{2}{*}{Parallel} & Perpen- & \multirow{2}{*}{Collinear} & Co- & \multirow{2}{*}{Same len.} & \multirow{2}{*}{Val. len.} & \multirow{2}{*}{Same $\angle$} & \multirow{2}{*}{Val. $\angle$} \\
 &  & -angle & -rilateral &  & -dicular &  & -circular &  &  &  &  \\
\midrule
\multirow{2}*{VL-Rethinker-7B} & Backbone & 69.1 & 52.3 & 49.0 & 78.2 & 59.5 & 65.6 & 57.4 & 86.6 & 50.0 & 68.1 \\
 & MM-RL & 67.6 & 55.2 & 60.6 & 76.6 & 67.4 & 81.7 & 50.0 & 86.9 & 50.0 & 75.8 \\
\midrule
\multirow{2}*{ShareVL-R1-7B} & Backbone & 69.1 & 52.3 & 49.0 & 78.2 & 59.5 & 65.6 & 57.4 & 86.6 & 50.0 & 68.1 \\
 & MM-RL & 71.4 & 57.7 & 62.5 & 78.2 & 66.0 & 88.9 & 57.4 & 86.4 & 69.2 & 72.4 \\
\midrule
\multirow{3}*{R1-OneVision-7B} & Backbone & 69.1 & 52.3 & 49.0 & 78.2 & 59.5 & 65.6 & 57.4 & 86.6 & 50.0 & 68.1 \\
 & MM-SFT & 67.6 & 48.1 & 56.7 & 67.4 & 53.2 & 60.0 & 48.5 & 76.7 & 46.2 & 63.5 \\
 & MM-RL & 71.4 & 54.4 & 59.6 & 72.2 & 61.1 & 59.4 & 55.9 & 79.1 & 61.5 & 66.9 \\
\midrule
\multirow{3}*{Vision-R1-7B} & Backbone & 69.1 & 52.3 & 49.0 & 78.2 & 59.5 & 65.6 & 57.4 & 86.6 & 50.0 & 68.1 \\
 & MM-SFT & 65.8 & 49.4 & 43.3 & 73.5 & 64.8 & 60.0 & 45.6 & 78.6 & 61.5 & 73.8 \\
 & MM-RL & 70.9 & 56.1 & 49.0 & 76.1 & 65.7 & 77.8 & 57.4 & 88.9 & 61.5 & 76.9 \\
\midrule
\multirow{3}*{Revisual-R1-7B} & Backbone & 69.1 & 52.3 & 49.0 & 78.2 & 59.5 & 65.6 & 57.4 & 86.6 & 50.0 & 68.1 \\
 & Text-SFT & 79.1 & 64.0 & 65.4 & 77.5 & 59.8 & 67.8 & 57.4 & 81.7 & 69.2 & 82.3 \\
 & MM-RL & 84.3 & 72.0 & 65.4 & 82.8 & 68.0 & 77.2 & 60.3 & 88.1 & 84.6 & 87.8 \\
\midrule
\multirow{3}*{OVR-7B} & Backbone & 69.1 & 52.3 & 49.0 & 78.2 & 59.5 & 65.6 & 57.4 & 86.6 & 50.0 & 68.1 \\
 & Text-SFT & 83.5 & 61.5 & 79.8 & 82.2 & 61.0 & 69.4 & 57.4 & 83.2 & 69.2 & 86.0 \\
 & MM-RL & 82.2 & 67.8 & 76.9 & 85.9 & 66.9 & 80.6 & 50.0 & 83.0 & 80.8 & 89.0 \\
\midrule
\multirow{2}*{GLM-4.1V-9B} & Backbone & 74.3 & 52.3 & 76.9 & 77.8 & 68.2 & 74.4 & 57.4 & 90.1 & 53.8 & 74.5 \\
 & MM-RL & 82.2 & 66.5 & 88.5 & 88.8 & 66.2 & 91.7 & 82.4 & 95.3 & 92.3 & 88.0 \\
\midrule
\multirow{2}*{VL-Rethinker-72B} & Backbone & 66.7 & 49.8 & 76.9 & 80.4 & 66.8 & 94.4 & 67.6 & 94.5 & 69.2 & 87.4 \\
 & MM-RL & 83.3 & 57.7 & 79.8 & 84.3 & 65.2 & 96.7 & 64.7 & 93.7 & 80.8 & 86.7 \\
\midrule
\multirow{2}*{OpenAI 4O / O3} & Backbone & 81.9 & 60.3 & 78.8 & 80.3 & 69.5 & 95.0 & 67.6 & 93.4 & 80.8 & 87.5 \\
 & Thinking & 94.0 & 95.4 & 95.2 & 92.8 & 94.1 & 93.9 & 95.6 & 98.9 & 100.0 & 96.7 \\
\midrule
\multirow{2}*{Gemini-2.5-Flash} & Backbone & 92.7 & 90.0 & 97.1 & 92.8 & 89.5 & 96.1 & 95.6 & 97.5 & 88.5 & 92.1 \\
 & Thinking & 93.2 & 92.1 & 97.1 & 95.5 & 91.8 & 96.7 & 98.5 & 98.4 & 92.3 & 93.9 \\
\midrule
\multirow{2}*{Claude-4-Sonnet} & Backbone & 88.6 & 75.3 & 79.8 & 85.2 & 82.3 & 95.0 & 77.9 & 98.1 & 84.6 & 92.5 \\
 & Thinking & 93.5 & 93.7 & 85.6 & 88.8 & 83.1 & 89.4 & 83.8 & 98.5 & 88.5 & 93.7 \\
\bottomrule
\end{tabular}}
\caption{Perception probe accuracy by question types.}
\label{tab:probe_perf}
\end{table*}

\paragraph{Perception probe results by question type.}
\Cref{tab:probe_perf} reports the full benchmark scores for each perception probe question type, providing the numerical values underlying~\Cref{fig:perception_category}.

\paragraph{Error type analysis.}
\Cref{fig:error_types} shows full error type analysis results for all models tested in this work.

\subsection{\DATANAMEG}
\label{subsec:ax_exp_general}

\begin{figure*}[ht]
\begin{center}
\includegraphics[width=0.8\textwidth]{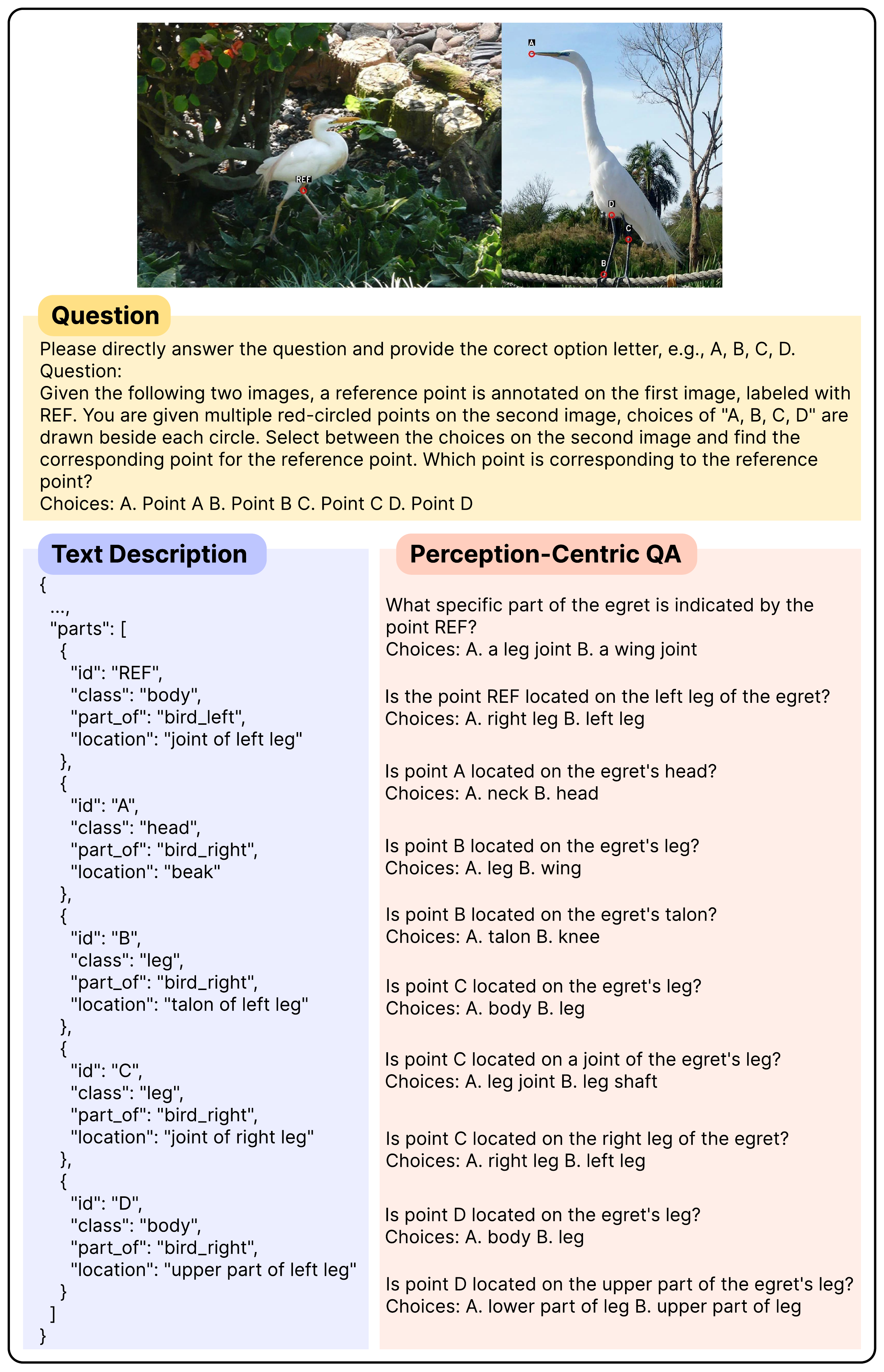}
\end{center}
\caption{\textbf{Data samples from \DATANAMEG.} We curate problem instances from a prior dataset~(\cite{fu2024blink}) and annotate the text description and perception-centric question-answers.}
\label{fig:ax_sample_general1}
\end{figure*}

\begin{figure*}[ht]
\begin{center}
\includegraphics[width=0.8\textwidth]{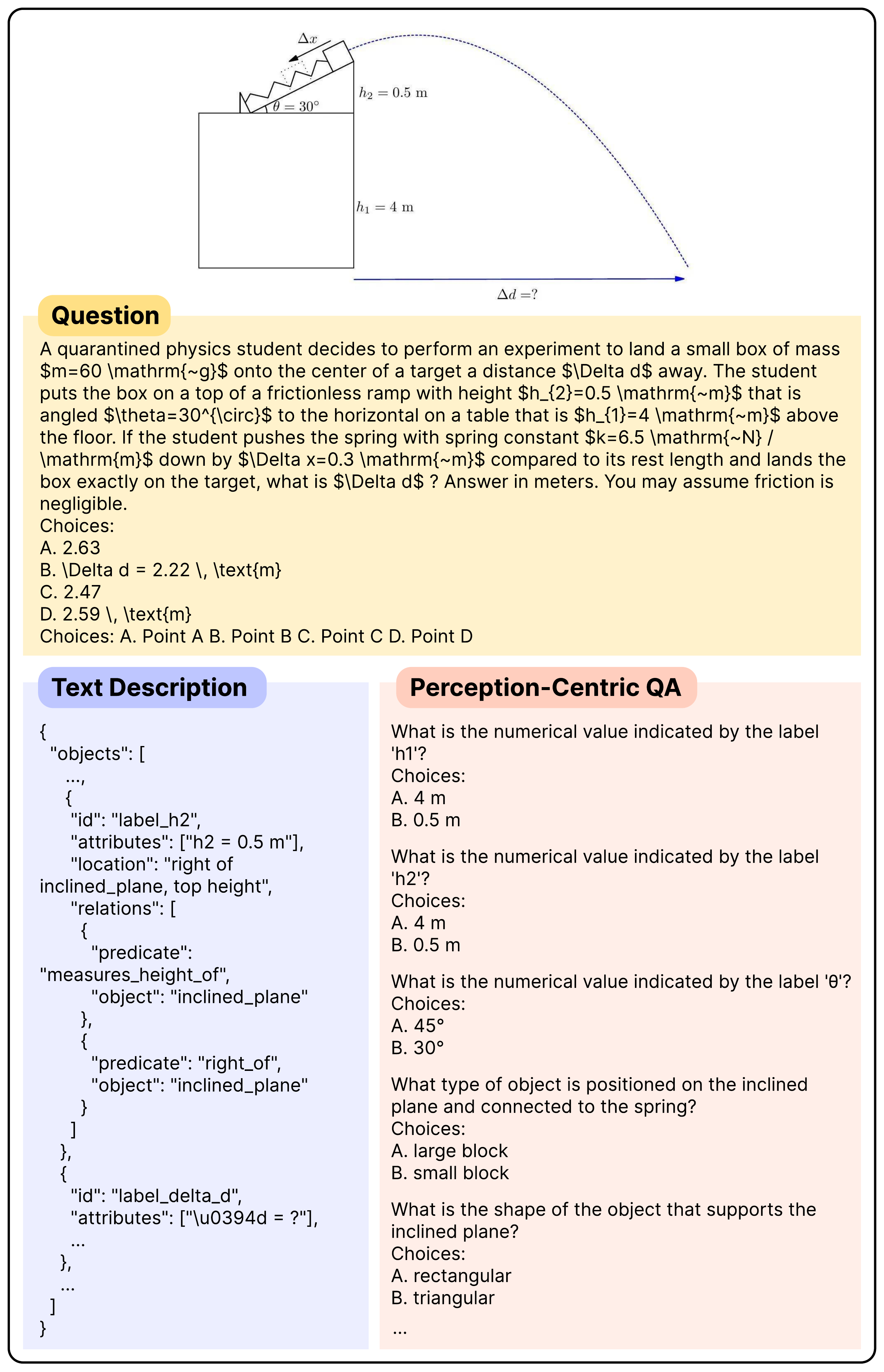}
\end{center}
\caption{\textbf{Data samples from \DATANAMEG.} We curate problem instances from a prior dataset~(\cite{hao2025emma}) and annotate the text description and perception-centric question-answers.}
\label{fig:ax_sample_general2}
\end{figure*}

\begin{figure*}[ht]
\begin{center}
\includegraphics[width=0.9\textwidth]{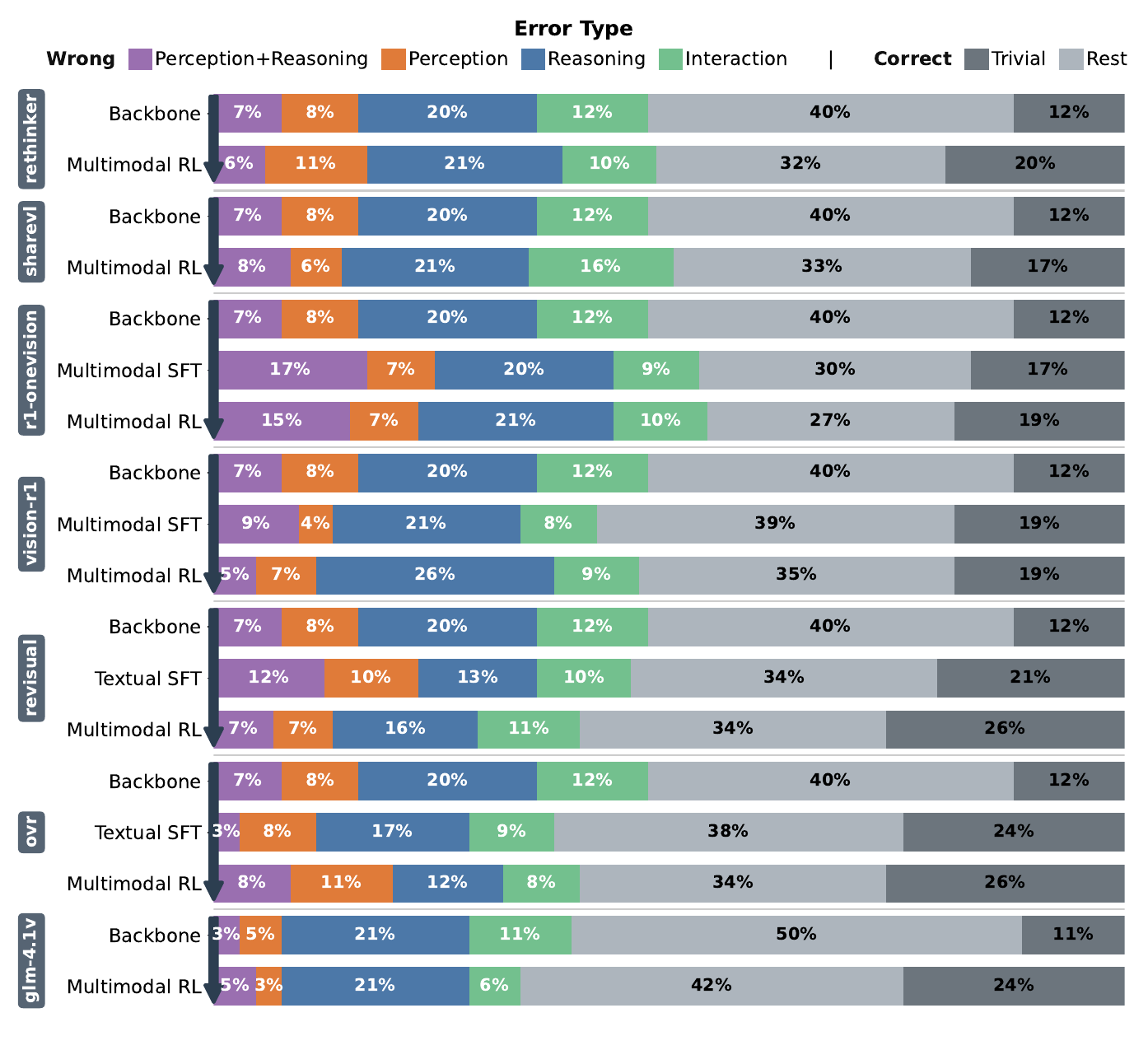}
\end{center}
\vspace{-15pt} %
\caption{\textbf{Error type distribution across models on \DATANAMEG.} Most shifts are associated with perception-related cases.}
\label{fig:error_types_general}
\end{figure*}

\Cref{fig:error_types_general} presents the error-type distribution for \DATANAMEG. Consistent with the main experiment (\Cref{subsubsec:exp_error_type}), most effects of multimodal reasoning finetuning concentrate on perception-related cases (\textit{Perception \& Reasoning} or \textit{Perception}). However, these gains are less stable, reflecting that \DATANAMEG spans broader domains than \DATANAME and often demands out-of-distribution generalization from the multimodal training sets. An exception is textual SFT models, which show substantial reductions in pure \textit{Reasoning} errors. This indicates that, unlike in math geometry tasks, the diverse reasoning skills required for \DATANAMEG are not well represented in the backbone (Qwen-2.5-VL). Finally, the higher fraction of \textit{Trivial} correct cases arises from \DATANAMEG's multiple-choice format, in contrast to the open-ended geometry subset \DATANAME.

\section{Qualitative Examples}
\label{sec:ax_examples}

\begin{figure*}[t]
\begin{center}
\includegraphics[width=0.8\textwidth]{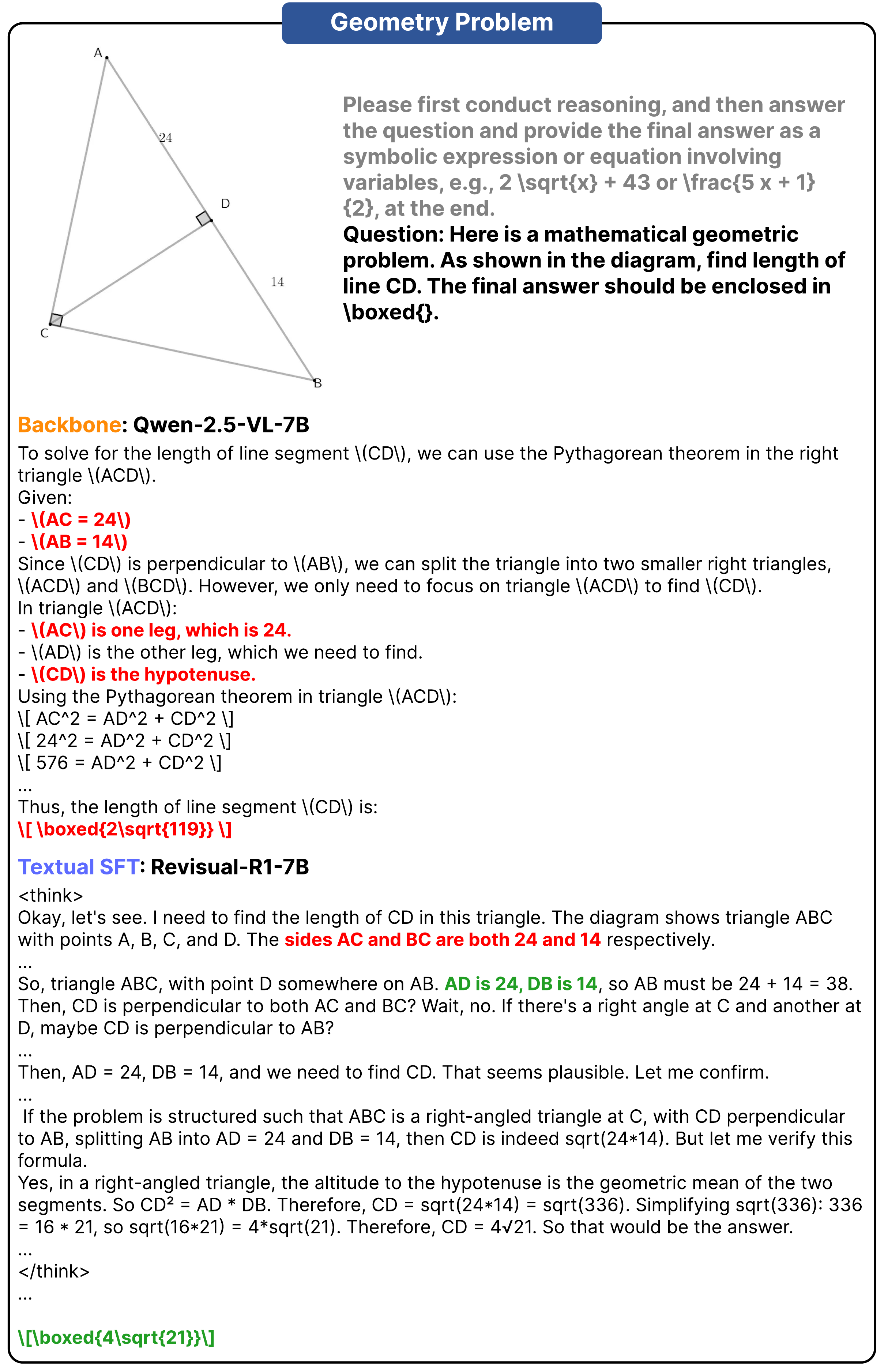}
\end{center}
\caption{\textbf{Model responses for a downstream geometry problem in \DATANAME.} The backbone model fails in perception, leading to an incorrect answer. The textual SFT model initially makes the same mistake but recovers the perceptual detail and produces the correct answer.}
\label{fig:ax_sample_changed1}
\end{figure*}

\begin{figure*}[t]
\begin{center}
\includegraphics[width=0.8\textwidth]{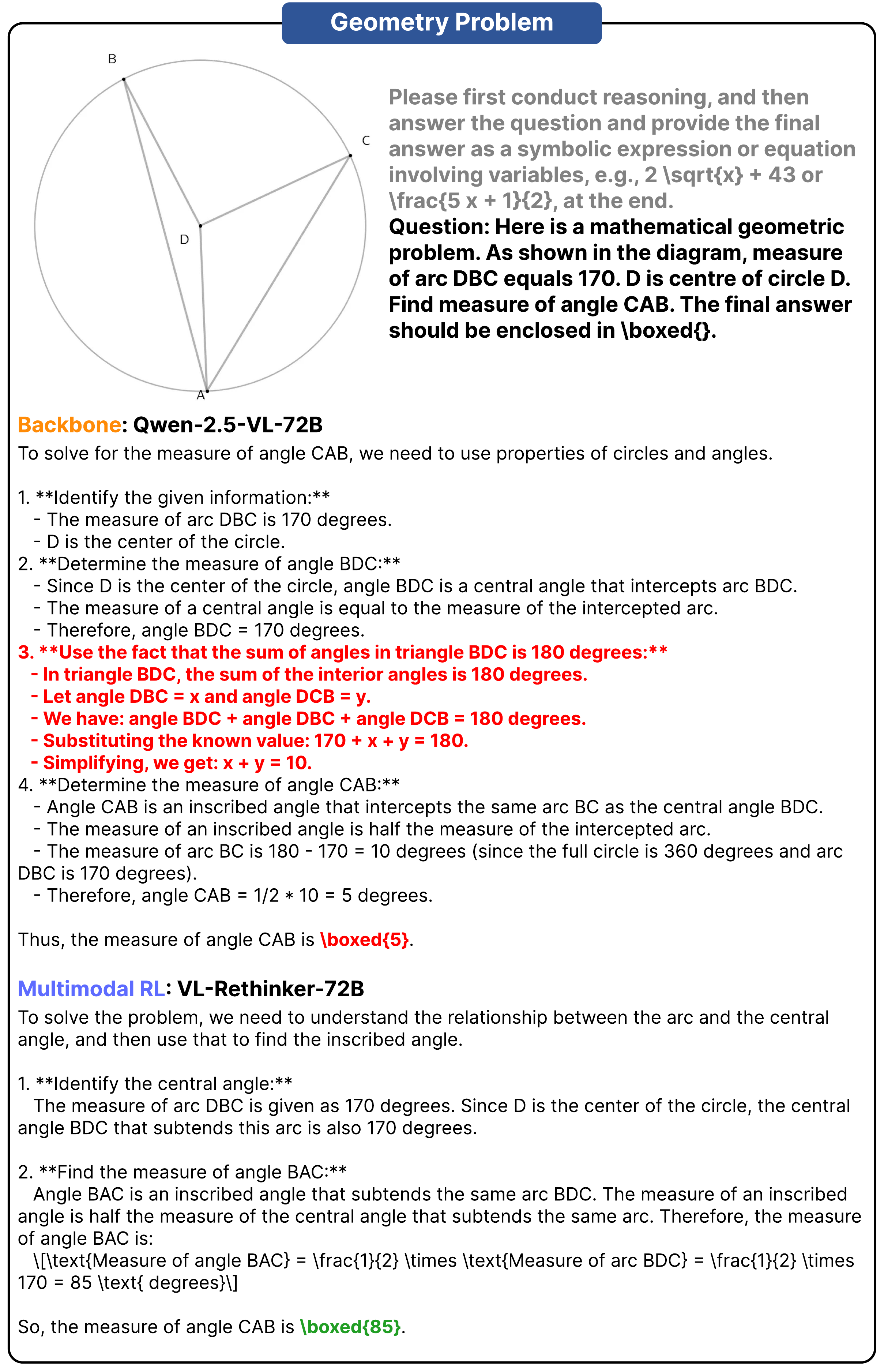}
\end{center}
\caption{\textbf{Model responses for a downstream geometry problem in \DATANAME.} The backbone model exhibits a reasoning error, whereas the multimodal RL model provides the correct solution.}
\label{fig:ax_sample_changed2}
\end{figure*}

\begin{figure*}[t]
\begin{center}
\includegraphics[width=0.8\textwidth]{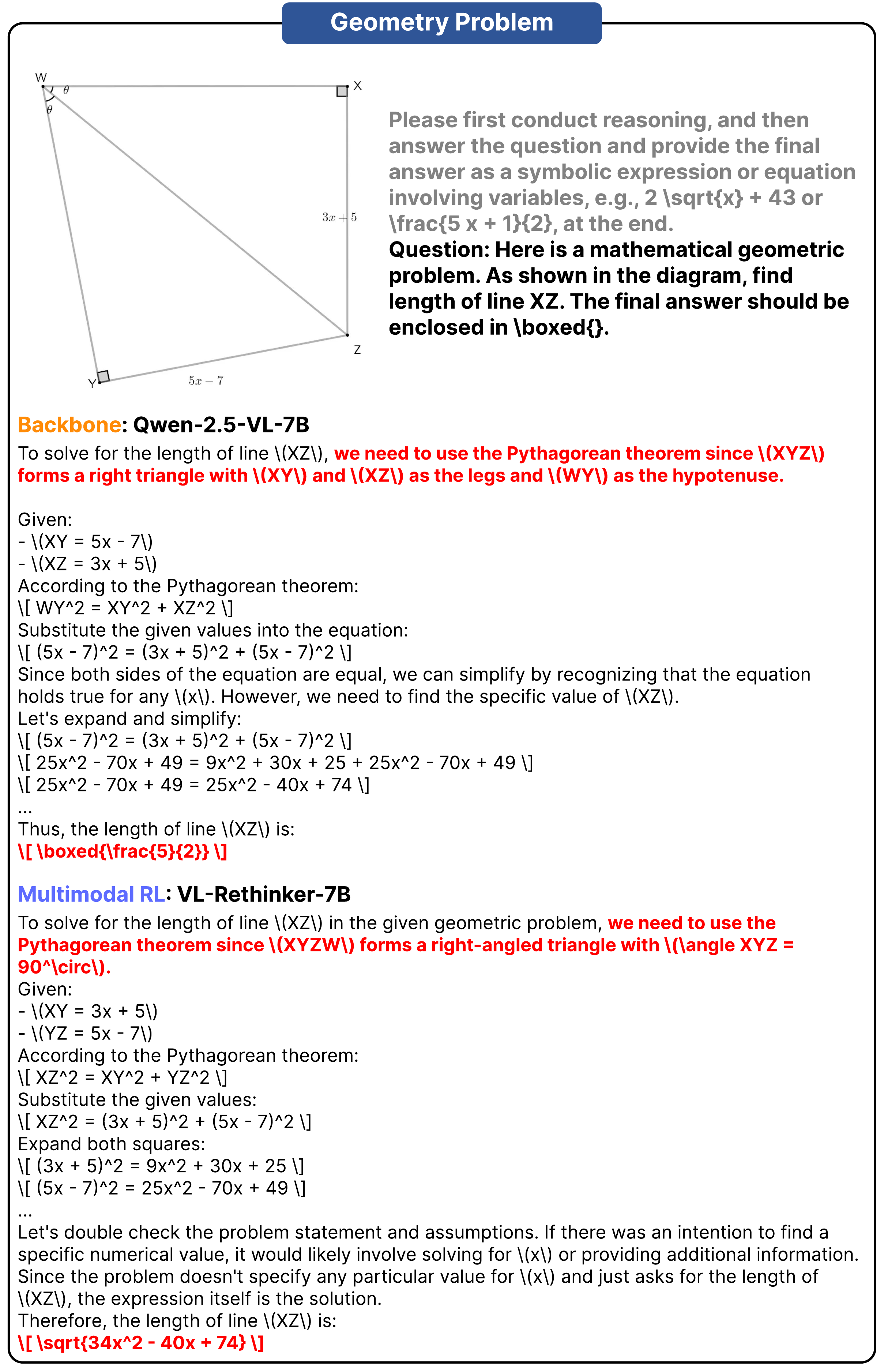}
\end{center}
\caption{\textbf{Model responses for a downstream geometry problem in \DATANAME.} The multimodal RL model inherits the backbone model's flawed reasoning, leading to perception errors and downstream failure.}
\label{fig:ax_sample_failure1}
\end{figure*}

\begin{figure*}[t]
\begin{center}
\includegraphics[width=0.8\textwidth]{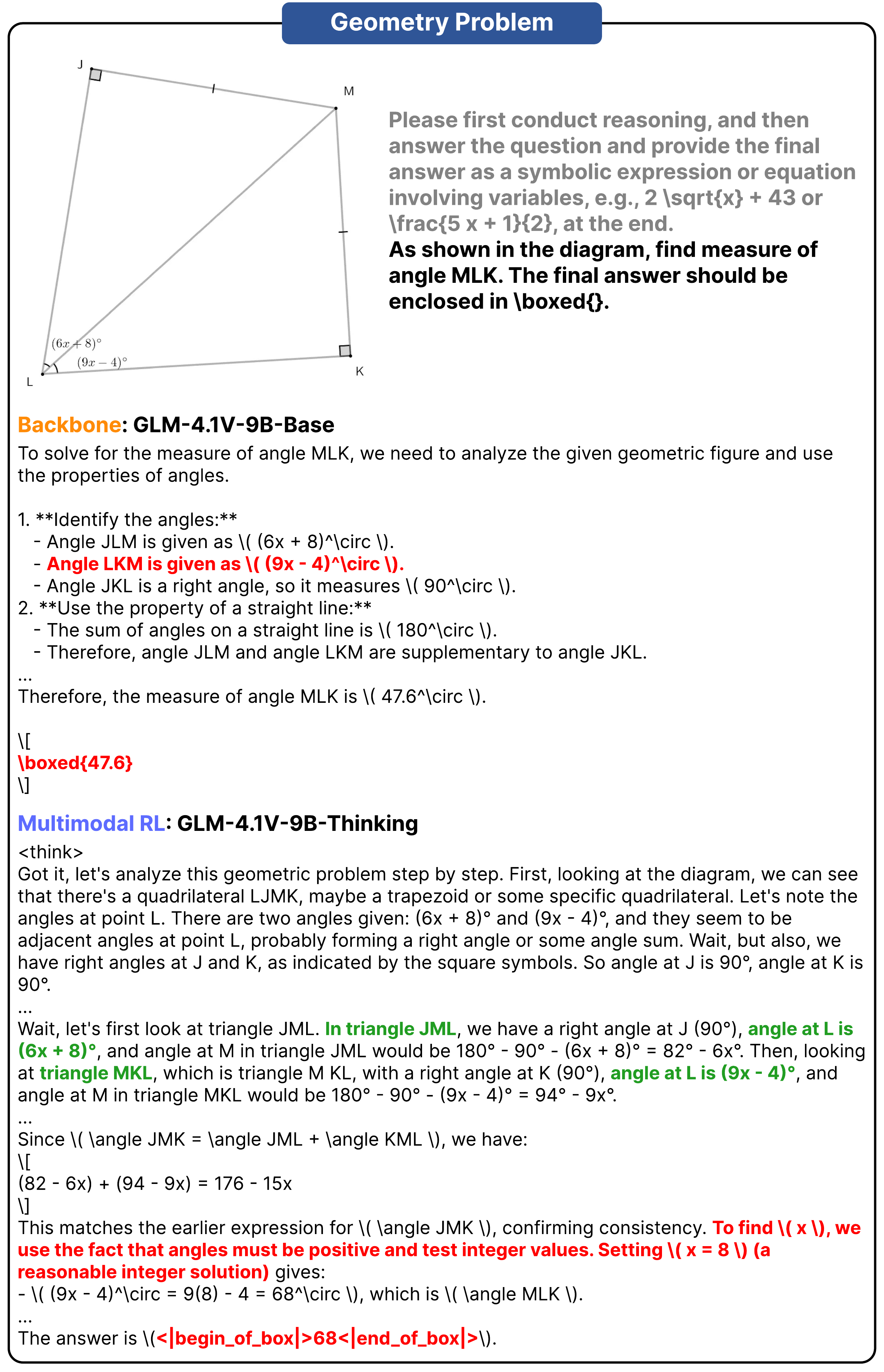}
\end{center}
\caption{\textbf{Model responses for a downstream geometry problem in \DATANAME.} The multimodal RL model corrects the perceptual error of the backbone model but ultimately fails to produce the correct answer due to a reasoning error.}
\label{fig:ax_sample_failure2}
\end{figure*}

\paragraph{Sample outputs on downstream geometry problems.}
We compare the finetuned multimodal reasoners with their corresponding backbone MLLMs.
\Cref{fig:ax_sample_changed1} and~\Cref{fig:ax_sample_changed2} present cases where finetuning corrected an initially wrong answer, while
\Cref{fig:ax_sample_failure1} and~\Cref{fig:ax_sample_failure2} illustrate cases where the model produced incorrect answers both before and after finetuning.

\begin{figure*}[t]
\begin{center}
\includegraphics[width=0.8\textwidth]{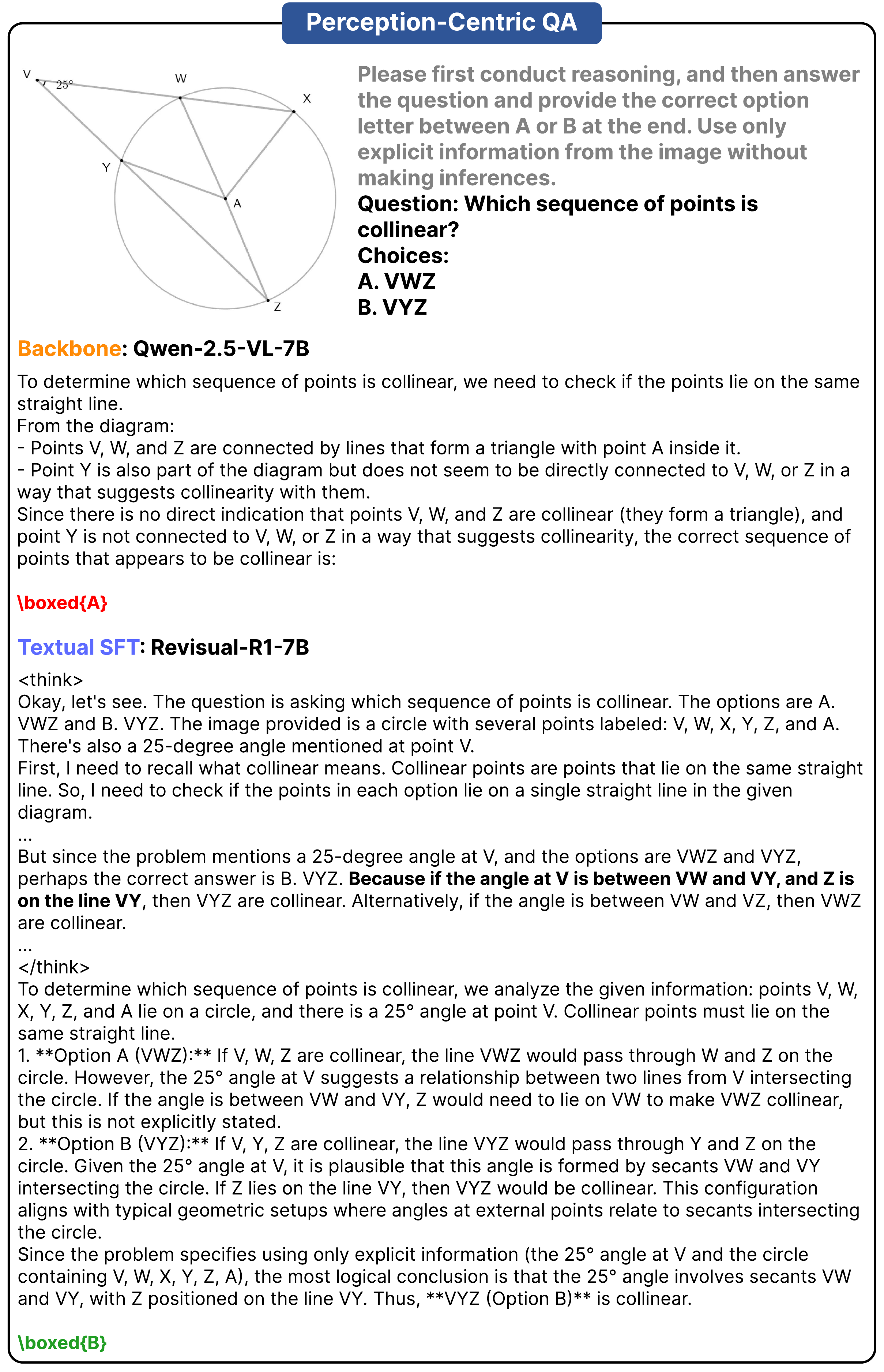}
\end{center}
\caption{\textbf{Model responses for a perception probe in \DATANAME.} The textual SFT model corrects the perceptual error present in the backbone model.}
\label{fig:ax_sample_perc1}
\end{figure*}

\begin{figure*}[t]
\begin{center}
\includegraphics[width=0.8\textwidth]{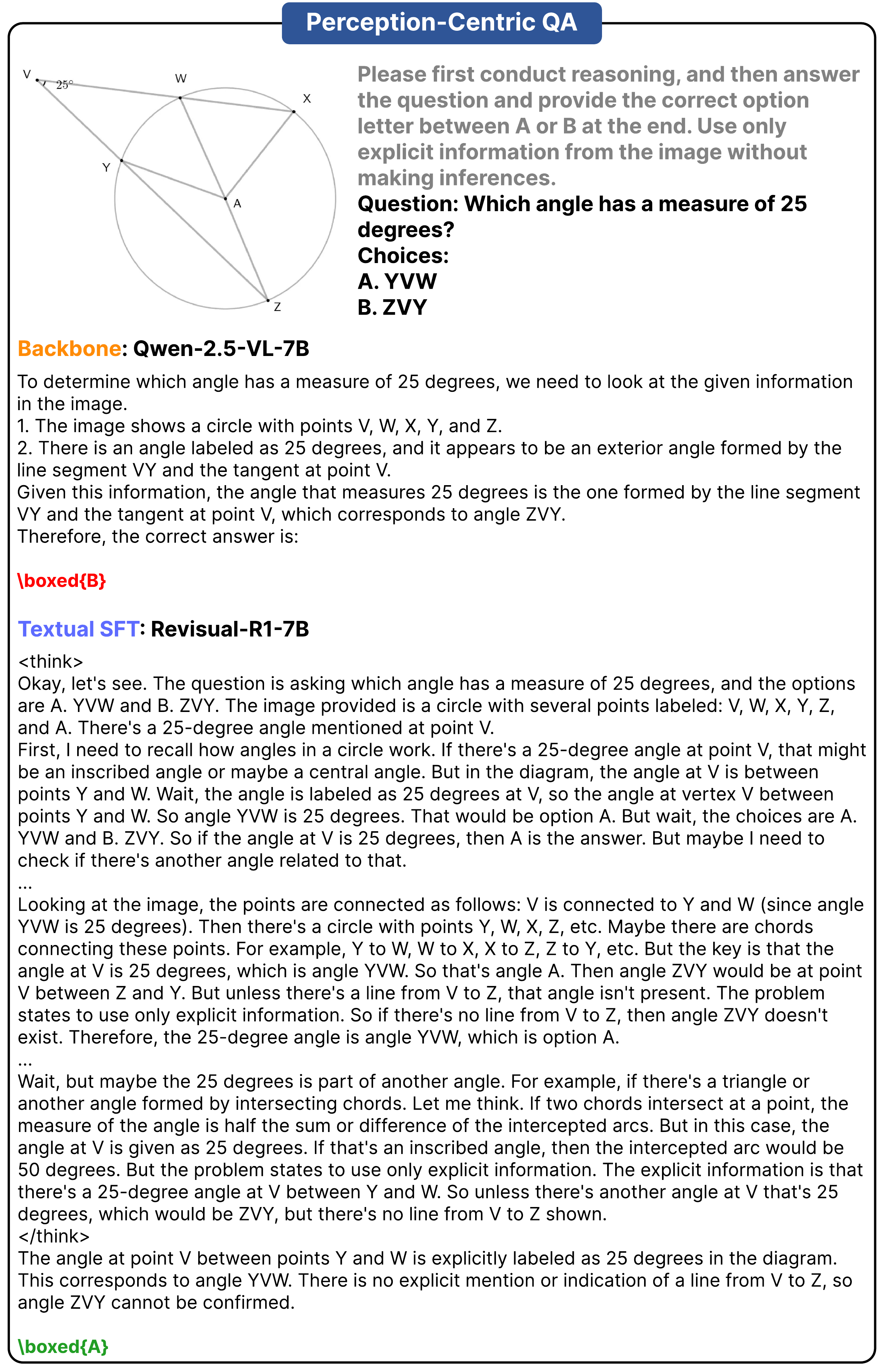}
\end{center}
\caption{\textbf{Model responses for a perception probe in \DATANAME.} The textual SFT model corrects the perceptual error present in the backbone model.}
\label{fig:ax_sample_perc2}
\end{figure*}

\begin{figure*}[t]
\begin{center}
\includegraphics[width=0.8\textwidth]{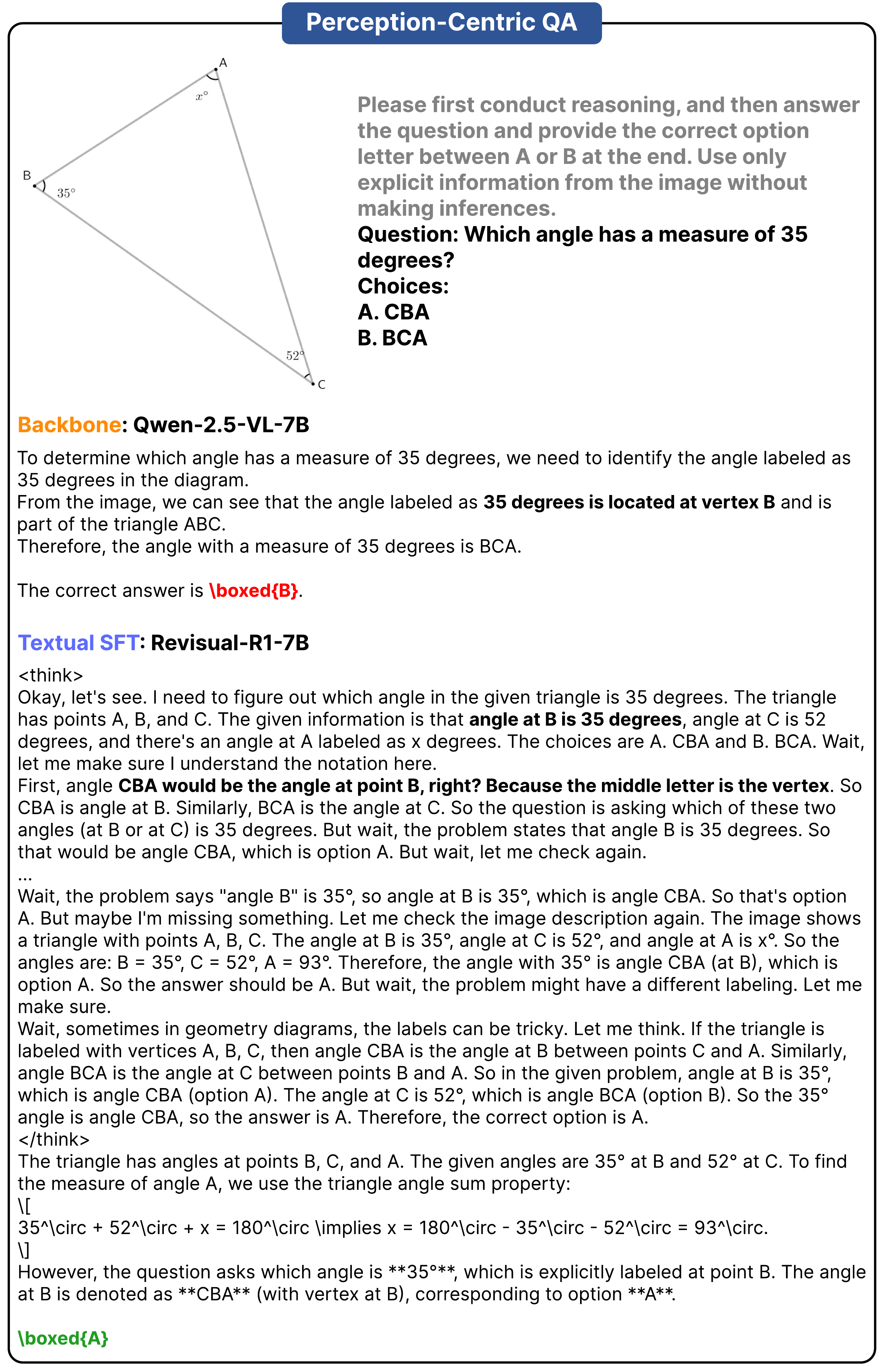}
\end{center}
\caption{\textbf{Model responses for a perception probe in \DATANAME.} The textual SFT model corrects the perceptual error present in the backbone model.}
\label{fig:ax_sample_perc3}
\end{figure*}

\paragraph{Sample outputs on perception probes.}
\Cref{fig:ax_sample_perc1},~\Cref{fig:ax_sample_perc2}, and~\Cref{fig:ax_sample_perc3} show cases where textual SFT corrected initially wrong answers. 
Consistent with the quantitative results in~\Cref{subsubsec:exp_perception}, the stronger cognitive patterns induced by textual SFT also promote improved perception.

\section{Broader Impact}
\label{sec:broader_impact}

Our goal is to improve the evaluation and understanding of multimodal reasoning models. By decomposing performance into perception, reasoning, and the residual multimodal-specific category, our benchmark provides more diagnostic insight than aggregate accuracy and helps clarify which components are responsible for observed gains.

Our work focuses on mathematical geometry problems and is constructed entirely from publicly available problem sources and synthetic derivations. It does not involve personal data, sensitive attributes, or application domains with direct societal deployment. As such, the benchmark primarily serves as a scientific analysis tool for the research community.

As with any benchmark, there is a risk that future work may over-optimize for its specific structure rather than broader generalization. To reduce this risk, the benchmark is designed with explicit subskill decompositions and controlled variants intended for diagnostic analysis rather than leaderboard-driven optimization. We do not anticipate additional ethical or societal risks beyond those commonly associated with advances in multimodal machine learning.

\section{Ethics Statement}
\label{sec:ethics}

\DATANAME consists solely of mathematical problems and does not involve human subjects or sensitive data. \DATANAMEG is constructed as an extension of existing benchmarks, and ethical considerations are therefore inherited from those sources. All human annotations were performed directly by the authors. As the benchmark is derived from publicly available resources, it does not raise additional privacy or copyright concerns beyond those already addressed in the original sources. As the content is limited to mathematical problems, concerns of bias, fairness, or harmful applications are not applicable.

\section{Reproducibility Statement}
\label{sec:repro}

All experiments are conducted using existing models without additional training. The complete list of models is provided in~\Cref{tab:model_hf_groups}, with hyperparameter configurations in~\Cref{sec:ax_impl}. Results are reported under deterministic decoding, aside from a small subset of models requiring random sampling to mitigate repetition, and API models, which use provider defaults (\Cref{sec:ax_impl}). Minor nondeterminism from computation kernels, common across current LLM decoding environments (both local and API-based), is generally not treated as a controlled factor~\citep{he2025nondeterminism}. As API services may evolve over time with undocumented changes, our main experiments focus on open-weight models, with API model results reported as auxiliary reference.

\clearpage
\section*{NeurIPS Paper Checklist}

\begin{enumerate}

\item {\bf Claims}
    \item[] Question: Do the main claims made in the abstract and introduction accurately reflect the paper's contributions and scope?
    \item[] Answer: \answerYes{}
    \item[] Justification: The abstract and~\Cref{sec:intro} list three contributions---a multi-axial evaluation framework that decomposes performance into perception, reasoning, and a residual multimodal-specific category; controlled findings about how training strategies (multimodal RL, multimodal SFT, textual SFT, and their combinations) shape these components; and the \DATANAME and \DATANAMEG resources. Each contribution is supported by experiments in~\Cref{sec:exp} (Findings 1--5) and the dataset construction in~\Cref{sec:data,sec:ax_data}.

\item {\bf Limitations}
    \item[] Question: Does the paper discuss the limitations of the work performed by the authors?
    \item[] Answer: \answerYes{}
    \item[] Justification: A dedicated Limitations section is provided in~\Cref{sec:ax_limitations} (pointed to from the end of~\Cref{sec:conclusion}), covering: (i) the geometry-domain scope of the controlled analyses, with \DATANAMEG used for broader (less formally controlled) coverage; and (ii) the comparison of publicly released models rather than controlled retraining. The geometry-scope point is expanded in~\Cref{subsec:data_general}.

\item {\bf Theory assumptions and proofs}
    \item[] Question: For each theoretical result, does the paper provide the full set of assumptions and a complete (and correct) proof?
    \item[] Answer: \answerNA{}
    \item[] Justification: The paper is empirical. The formalization in~\Cref{subsec:data_form} introduces notation for semantic states, query operators, and perception probes to specify the evaluation protocol; it states no theorems and makes no formal claims requiring proof.

    \item {\bf Experimental result reproducibility}
    \item[] Question: Does the paper fully disclose all the information needed to reproduce the main experimental results?
    \item[] Answer: \answerYes{}
    \item[] Justification: All evaluated models with HuggingFace identifiers and decoding settings are listed in~\Cref{tab:model_hf_groups} and~\Cref{sec:ax_impl}. The dataset construction pipeline (semantic predicates, constraint solving with SLSQP, diagram rendering, modification operators, perception-probe templating) is detailed in~\Cref{subsec:data_gen,sec:ax_data}. Statistical methodology (paired bootstrap CIs over problem instances) is described in~\Cref{sec:exp}.

\item {\bf Open access to data and code}
    \item[] Question: Does the paper provide open access to the data and code?
    \item[] Answer: \answerNo{}
    \item[] Justification: We are preparing to release the dataset (\DATANAME, \DATANAMEG), the data-generation pipeline, and the evaluation scripts.

\item {\bf Experimental setting/details}
    \item[] Question: Does the paper specify all the training and test details?
    \item[] Answer: \answerYes{}
    \item[] Justification: This work performs no additional training; it evaluates released models. Evaluation details---hardware (4$\times$A100 80GB for 7--9B models, 8$\times$ for 72B), inference framework (Eureka ML Insights with vLLM), decoding configuration (greedy by default; temperature 0.6 / top-$p$ 0.65 for models with text degeneration), and maximum generation length (32{,}768 tokens, extended to 48{,}000 for OVR)---are reported in~\Cref{sec:ax_impl}. Per-model recommended decoding settings are followed as noted in~\Cref{sec:exp}.

\item {\bf Experiment statistical significance}
    \item[] Question: Does the paper report error bars or statistical significance?
    \item[] Answer: \answerYes{}
    \item[] Justification: As stated in~\Cref{sec:exp}, paired bootstrap confidence intervals over problem instances are reported for all pre-/post-finetuning deltas (\Cref{fig:text,fig:perception,fig:diagram_ood}). Decoding is deterministic for most models, so the intervals reflect benchmark sampling variability rather than decoding stochasticity. Raw accuracies and diagnostic visualizations (\Cref{fig:main,fig:robustness,fig:perception_category}) are reported descriptively, with full numerical tables in~\Cref{tab:ax_main_perf,tab:probe_perf}.

\item {\bf Experiments compute resources}
    \item[] Question: Does the paper provide sufficient information on compute resources?
    \item[] Answer: \answerYes{}
    \item[] Justification: \Cref{sec:ax_impl} reports that 7--9B models are evaluated on 4$\times$NVIDIA A100 80GB GPUs and 72B models on 8$\times$A100 80GB, with vLLM-accelerated inference.

\item {\bf Code of ethics}
    \item[] Question: Does the research conform with the NeurIPS Code of Ethics?
    \item[] Answer: \answerYes{}
    \item[] Justification: \DATANAME consists of mathematical geometry problems derived from FormalGeo-7K and contains no personal, demographic, or sensitive data. \DATANAMEG re-curates publicly released multimodal benchmarks under their original terms. All manual filtering and validation were conducted by the authors. The work poses no foreseeable harms and conforms with the NeurIPS Code of Ethics.

\item {\bf Broader impacts}
    \item[] Question: Does the paper discuss broader impacts?
    \item[] Answer: \answerYes{}
    \item[] Justification: A dedicated Broader Impact statement appears in~\Cref{sec:broader_impact} and the Ethics Statement in~\Cref{sec:ethics}. Together they describe the intended scientific use (diagnostic evaluation that redirects research away from aggregate-accuracy optimization), the absence of personal or sensitive content, and the risk of leaderboard-style over-optimization that motivates the benchmark's subskill decomposition.

\item {\bf Safeguards}
    \item[] Question: Does the paper describe safeguards for responsible release?
    \item[] Answer: \answerNA{}
    \item[] Justification: We are preparing to release an evaluation benchmark (geometry problems, diagrams, textual descriptions, and perception probes) and per-model evaluation outputs. It releases no pretrained models, generative systems, or scraped data with privacy or misuse risk, so dedicated safeguards (gated access, content filters, etc.) are not applicable.

\item {\bf Licenses for existing assets}
    \item[] Question: Are existing assets properly credited?
    \item[] Answer: \answerYes{}
    \item[] Justification: All source datasets, evaluated models, and software dependencies are cited at their points of use. License terms for the assets we redistribute or build upon are as follows:
    \begin{itemize}[leftmargin=*,nosep]
        \item FormalGeo-7K~\citep{zhang2023formalgeo}: MIT License.
        \item BLINK~\citep{fu2024blink}: Apache-2.0.
        \item MMMU-Pro~\citep{yue2024mmmupro}: Apache-2.0.
        \item MathVista~\citep{lu2024mathvista}: CC-BY-SA-4.0; the dataset license permits use as a test set but prohibits use as training data, which we comply with (we evaluate on it and use it only as a source pool for \DATANAMEG curation).
        \item V*Bench~\citep{wu2024vstar}: codebase released under MIT License; the benchmark distribution does not state a separate dataset license, so we follow the project terms.
        \item SpatialEval~\citep{wang2024spatialeval} and EMMA~\citep{hao2025emma}: no explicit license is stated by the authors at their official distributions; we use them under standard academic-research norms in line with their public release.
        \item Software: SciPy~\citep{virtanen2020scipy} (BSD-3-Clause), Matplotlib~\citep{hunter2007matplotlib} (PSF-based), vLLM~\citep{kwon2023efficient} (Apache-2.0), and Eureka ML Insights~\citep{balachandran2024eureka} (MIT).
        \item Evaluated models: the open-weight models in~\Cref{tab:model_hf_groups} are used under the licenses set by their respective HuggingFace pages; closed-weight API models (OpenAI, Gemini, Claude) are accessed under the providers' terms of service.
        \item Visual icons used in~\Cref{fig:data_main} are credited to \texttt{flaticon.com}.
    \end{itemize}

\item {\bf New assets}
    \item[] Question: Are new assets well documented?
    \item[] Answer: \answerYes{}
    \item[] Justification: The new artifacts---\DATANAME (926 geometry problems with eight diagram variants and $\sim$7.03 perception probes per problem) and \DATANAMEG (107 cross-domain problems with $\sim$7.96 probes per problem)---are documented in~\Cref{sec:data,sec:ax_data}, including the symbolic specification, rendering pipeline, modification operators, probe templating, manual filtering, and per-subset statistics. Qualitative samples are provided in~\Cref{fig:ax_sample_general1,fig:ax_sample_general2,sec:ax_examples}. A datasheet, distribution URL, and final license terms (compatible with the upstream FormalGeo-7K MIT and source-benchmark licenses listed above) will be prepared and released alongside the dataset.

\item {\bf Crowdsourcing and research with human subjects}
    \item[] Question: Does the paper include details about crowdsourcing or human subjects?
    \item[] Answer: \answerNA{}
    \item[] Justification: No crowdworkers or external human subjects were involved. Manual filtering of rendered diagrams (\Cref{subsec:data_gen,sec:ax_data}) and validity verification of \DATANAMEG candidates (\Cref{subsec:ax_data_detail}) were performed directly by the authors.

\item {\bf Institutional review board (IRB) approvals}
    \item[] Question: Were IRB approvals obtained?
    \item[] Answer: \answerNA{}
    \item[] Justification: The work involves no human-subjects research, no participant data collection, and no interaction with external annotators, so IRB review is not applicable.

\item {\bf Declaration of LLM usage}
    \item[] Question: Does the paper describe the usage of LLMs if it is an important, original, or non-standard component of the core methods in this research? Note that if the LLM is used only for writing, editing, or formatting purposes and does \emph{not} impact the core methodology, scientific rigor, or originality of the research, declaration is not required.
    \item[] Answer: \answerYes{}
    \item[] Justification: As declared in~\Cref{sec:ax_impl}, large language models (ChatGPT GPT-4/5 class and Claude 4 Sonnet) were used to refine writing style and clarity and to surface candidate related work for manual screening. They were not used as part of the experimental methodology. The \DATANAMEG curation pipeline does use Gemini-2.5-Flash for filtering candidate problems (\Cref{subsec:ax_data_detail}); this usage is methodological and is fully described.

\end{enumerate}

\end{document}